\documentclass[10pt,journal,compsoc]{IEEEtran}

\ifCLASSOPTIONcompsoc
  \usepackage[nocompress]{cite}
\else
  \usepackage{cite}
\fi
\ifCLASSINFOpdf
  \usepackage[pdftex]{graphicx} 
\else
\fi
\usepackage{amsmath, amssymb}
\ifCLASSOPTIONcompsoc
 \usepackage[caption=false,font=footnotesize,labelfont=sf,textfont=sf]{subfig}
\else
 \usepackage[caption=false,font=footnotesize]{subfig}
\fi
\usepackage{url}
\usepackage{tikz}
\usepackage{pgfplots}
\pgfplotsset{compat=newest}
\usetikzlibrary{plotmarks}
\usepackage{filecontents}
\usepackage{hhline}

\makeatletter
\pgfplotsset{
    /tikz/max node/.style={
        anchor=south, color=red
    },
    /tikz/min node/.style={
        anchor=north
    },
    mark min/.style={
        point meta rel=per plot,
        visualization depends on={x \as \xvalue},
        scatter/@pre marker code/.code={%
            \ifx\pgfplotspointmeta\pgfplots@metamin
                \def\markopts{}%
                \node [min node] {
                    \pgfmathprintnumber[fixed]{\xvalue},%
                    \pgfmathprintnumber[fixed]{\pgfplotspointmeta}
                };
            \else
                \def\markopts{mark=none}
            \fi
            \expandafter\scope\expandafter[\markopts,every node near coord/.style=green]
        },%
        scatter/@post marker code/.code={%
            \endscope
        },
        scatter,
    },
    mark max/.style={
        point meta rel=per plot,
        visualization depends on={x \as \xvalue},
        scatter/@pre marker code/.code={%
        \ifx\pgfplotspointmeta\pgfplots@metamax
            \def\markopts{fill=red,draw=red}%
            \node [max node] {
                \pgfmathprintnumber[fixed]{\pgfplotspointmeta}
            };
        \else
            \def\markopts{mark=*}
        \fi
            \expandafter\scope\expandafter[\markopts]
        },%
        scatter/@post marker code/.code={%
            \endscope
        },
        scatter
    }
}
\makeatother

\newcommand{\ve}[1]{\mathbf{#1}} 
\newcommand{\ma}[1]{\mathrm{#1}} 

\begin{document}
%
\title{Hierarchical binary CNNs for landmark localization with limited resources}
%
%
%
%

\author{Adrian Bulat and Georgios Tzimiropoulos
    \IEEEcompsocitemizethanks{\IEEEcompsocthanksitem A. Bulat and G. Tzimiropoulos are with the School of Computer Science, University of Nottingham.\protect\\
        E-mail: \{adrian.bulat, yorgos.tzimiropoulos\}@nottingham.ac.uk}
    \thanks{Manuscript received April 19, 2005; revised August 26, 2015.}}

%
%

\markboth{Journal of \LaTeX\ Class Files,~Vol.~14, No.~8, August~2015}%
{Shell \MakeLowercase{\textit{et al.}}: Bare Advanced Demo of IEEEtran.cls for IEEE Computer Society Journals}
%



\IEEEtitleabstractindextext{%
    \begin{abstract}
        Our goal is to design architectures that retain the groundbreaking performance of Convolutional Neural Networks (CNNs) for landmark localization and at the same time are lightweight, compact and suitable for applications with limited computational resources. To this end, we make the following contributions: (a) we are the first to study the effect of neural network binarization on localization tasks, namely human pose estimation and face alignment. We exhaustively evaluate various design choices, identify performance bottlenecks, and more importantly propose multiple orthogonal ways to boost performance. (b) Based on our analysis, we propose a novel hierarchical, parallel and multi-scale residual architecture that yields large performance improvement over the standard bottleneck block while having the same number of parameters, thus bridging the gap between the original network and its binarized counterpart. (c) We perform a large number of ablation studies that shed light on the properties and the performance of the proposed block. (d) We present results for experiments on the most challenging datasets for human pose estimation and face alignment, reporting in many cases state-of-the-art performance. (e) We further provide additional results for the problem of facial part segmentation. Code can be downloaded from \url{https://www.adrianbulat.com/binary-cnn-landmarks}
    \end{abstract}

    }

\maketitle

\IEEEdisplaynontitleabstractindextext

%
\IEEEpeerreviewmaketitle

\ifCLASSOPTIONcompsoc
    \IEEEraisesectionheading{\section{Introduction}\label{sec:introduction}}
\else
    \section{Introduction}
	\label{sec:introduction}
\fi

%
%
%
%

\IEEEPARstart{T}{his} work is on localizing a predefined set of fiducial points on objects of interest which can typically undergo non-rigid deformations like the human body or face. Very recently, work based on Convolutional Neural Networks (CNNs) has revolutionized landmark localization, demonstrating results of remarkable accuracy even on the most challenging datasets for human pose estimation \cite{ bulat2016human,newell2016stacked,wei2016convolutional} and face alignment \cite{bulat2016two}. However, deploying (and training) such methods is computationally expensive, requiring one or more high-end GPUs, while the learned models typically require hundreds of MBs, thus rendering them completely unsuitable for real-time or mobile applications. This work is on highly accurate and robust yet efficient and lightweight landmark localization using binarized CNNs.

Our work is inspired by recent results of binarized CNN architectures on image classification \cite{rastegari2016xnor, courbariaux2016binarized}. Contrary to these works, we are the first to study the effect of neural network binarization on fine-grained tasks like landmark localization. Similarly to \cite{rastegari2016xnor, courbariaux2016binarized}, we find that binarization results in performance drop, however to address this we opted to investigate and propose several architectural innovations which led to the introduction of a novel hierarchical, parallel and multi-scale residual block, as opposed to investigating ways to improve the binarization process as proposed in \cite{rastegari2016xnor, courbariaux2016binarized}.  In summary, our main methodological contributions are:
\begin{figure}[!htb]
    \centering
    \subfloat[original]{\includegraphics[height=2.0in,trim={0.5cm 0.5cm 0.5cm 0.5cm},clip]{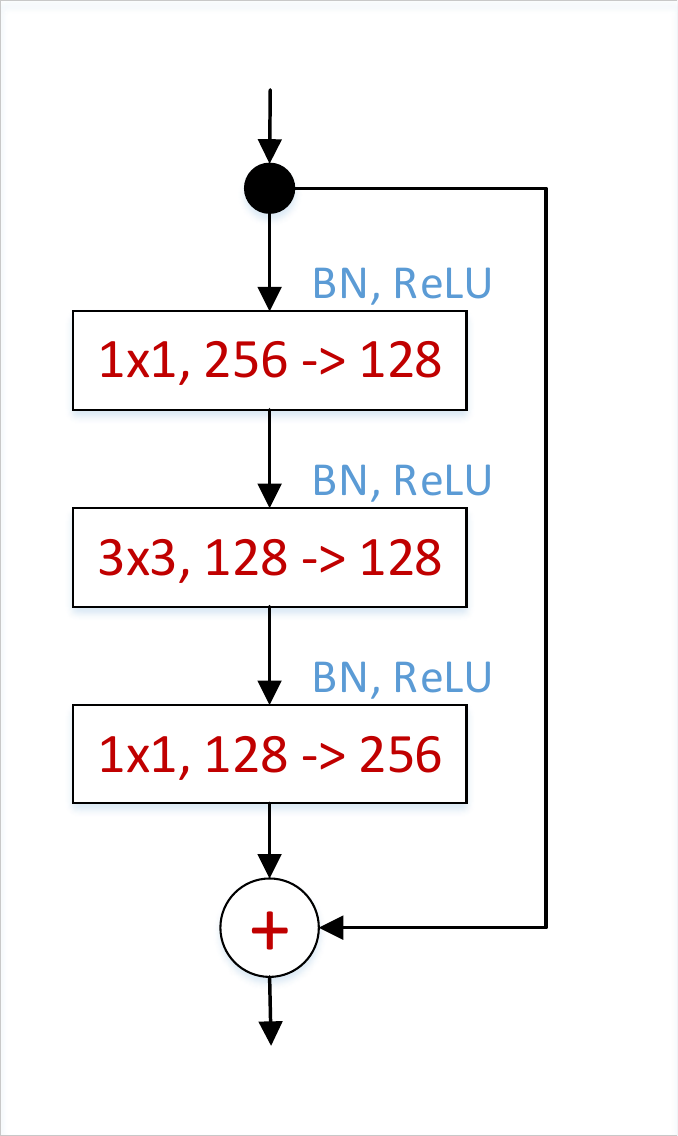}
        \label{fig:layers_original_intro}}
    \subfloat[proposed]{\includegraphics[height=2.0in,trim={0.5cm 0.5cm 0.5cm 0.5cm},clip]{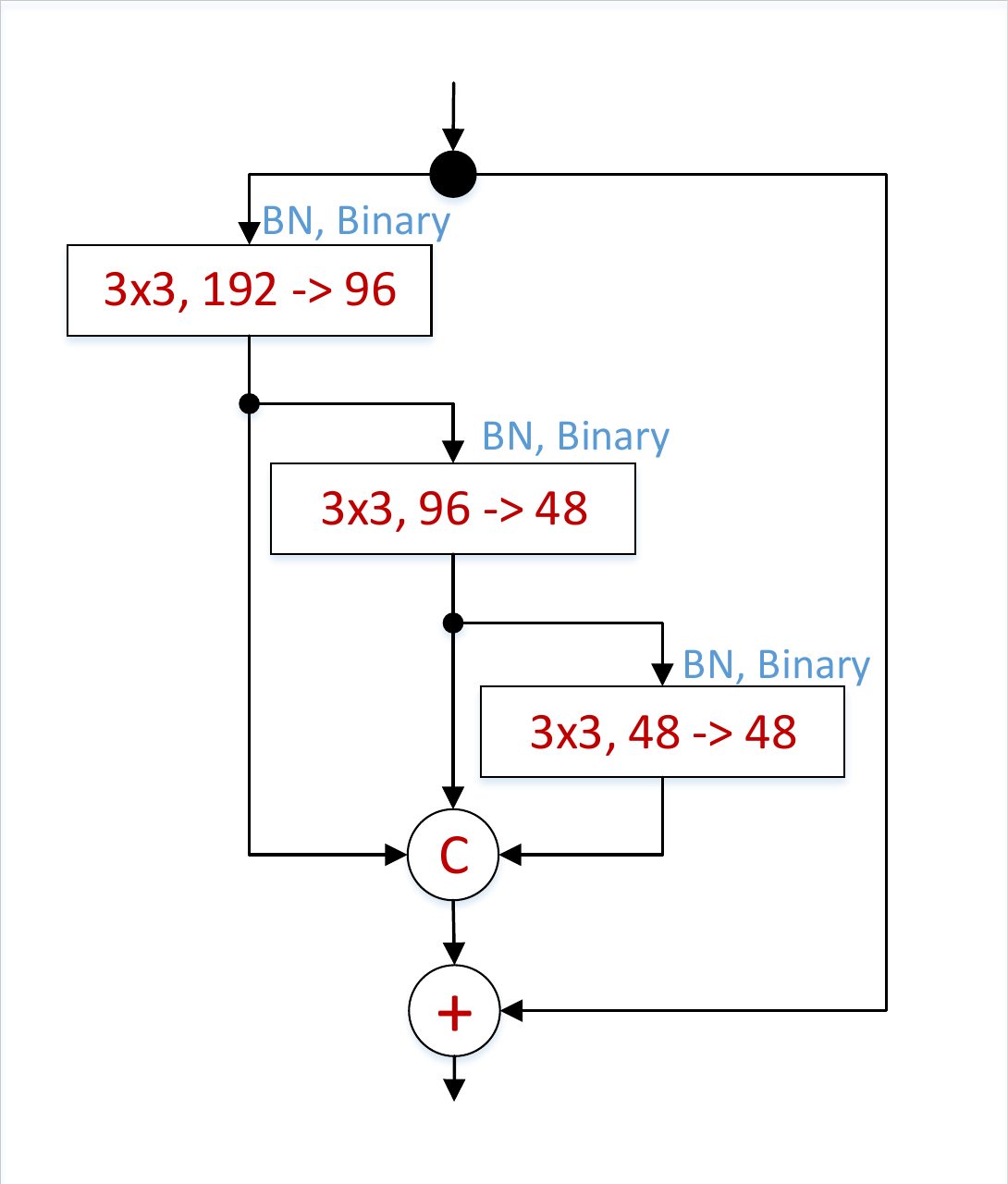}
        \label{fig:layers_wider_intro}}

    \caption{ (a) The original bottleneck layer of \cite{he2016identity}. (b) The proposed hierarchical parallel \& multi-scale structure: our block increases the receptive field size, improves gradient flow, is specifically designed to have (almost) the same number of parameters as the original bottleneck, does not contain $1 \times 1$ convolutions, and in general is derived from the perspective of improving the performance and efficiency for binary networks. \textbf{Note:} a layer is depicted as a rectangular block containing: its filter size, the number of input and output channels; "C" - denotes concatenation and "+" an element-wise sum.}
    \label{fig:intro_layers}
\end{figure}

\begin{itemize}
    \item
          We are the first to study the effect of binarization on state-of-the-art CNN architectures for the problem of localization, namely human pose estimation and face alignment. To this end, we exhaustively evaluate various design choices, and identify performance bottlenecks. More importantly, we describe multiple orthogonal ways to boost performance; see Subsections~\ref{ssec:wide_block},~\ref{ssec:size_filters} and~\ref{ssec:conv1}.
    \item
          Based on our analysis, we propose a new hierarchical, parallel and multi-scale residual architecture (see Subsection \ref{ssec:better}) specifically designed to work well for the binary case. Our block results in large performance improvement over the baseline binary residual block of \cite{he2016identity} (about 6\% in absolute terms when the same number of parameters are used (see Subsection \ref{sec:binary}, Table~\ref{tab:method_results})). {\color{black} Fig. \ref{fig:intro_layers} provides a comparison between the baseline residual block of \cite{he2016identity} and the one proposed in this work.} 
    \item 
         We investigate the effectiveness of more advanced extensions of the proposed block (see Section \ref{sec:advanced-block-architectures}) and improved network architectures  including network stacking (see Section \ref{sec:network-architectures}).  
\end{itemize}
Further experimental contributions include:
\begin{itemize}
    \item
          While our newly proposed block was developed with the goal of improving the performance of binary networks, we also show that the performance boost offered by the proposed architecture also generalizes to some extent for the case of real-valued networks (see Subsection~\ref{sec:real}).
    \item
          We perform a large number of ablation studies that shed light on the properties and the performance of the proposed block (see Sections~\ref{sec:binaryvsreal} and~\ref{sec:results}).
    \item
          We present results for experiments on the most challenging datasets for human pose estimation and face alignment, reporting in many cases state-of-the-art performance (see Section~\ref{sec:results}).
    \item
    	We further provide additional results for the problem of facial part segmentation (see Section \ref{sec:additional-experiments}).
\end{itemize}

Compared to our previous work in \cite{bulat2017binarized}, this paper investigates the effectiveness of more advanced binary architectures (both at block and network level), provides a more in-depth analysis of the proposed methods and results (including more qualitative ones) and additionally includes the aforementioned experiment on facial part segmentation.  






\section{Closely Related Work}\label{sec:related}

This Section reviews related work on network quantization, network design, and gives an overview of the state-of-the-art on human pose estimation and face alignment.

\subsection{Network quantization} Prior work \cite{holi1993finite} suggests that high precision parameters are not essential for obtaining top results for image classification. In light of this, \cite{courbariaux2014training, lin2015fixed} propose 16- and 8-bit quantization, showing negligible performance drop on a few small datasets \cite{krizhevsky2009learning}. \cite{zhou2016dorefa} proposes a technique which allocates different numbers of bits (1-2-6) for the network parameters, activations and gradients.

Binarization (i.e. the extreme case of quantization) was long considered to be impractical due to the destructive property of such a representation \cite{courbariaux2014training}. Recently \cite{soudry2014expectation} showed this not to be the case and that by quantizing to $\{-1,1\}$ good results can be actually obtained. \cite{courbariaux2015binaryconnect} introduces a new technique for training CNNs that uses binary weights for both forward and backward passes, however, the real parameters are still required during training.  The work of \cite{courbariaux2016binarized} goes one step further and binarizes both parameters and activations. In this case multiplications can be replaced with elementary binary operations \cite{courbariaux2016binarized}. By estimating the binary weights with the help of a scaling factor, \cite{rastegari2016xnor} is the first work to report good results on a large dataset (ImageNet). Notably, our method makes use of the recent findings from \cite{rastegari2016xnor} and \cite{courbariaux2016binarized} using the same way of quantizing the weights and replacing multiplications with bit-wise \textit{xor} operations.

Our method differs from all aforementioned works in two key respects: (a) instead of focusing on image classification, we are the first to study neural network binarization in the context of a fine-grained computer vision task namely landmark localization (human pose estimation and facial alignment) by predicting a dense output (heatmaps) in a fully convolutional manner, and (b) instead of enhancing the results by improving the quantization method, we follow a completely different path, by enhancing the performance via proposing a novel architectural design for a hierarchical, parallel and multi-scale residual block.

\subsection{Block design} The proposed method uses a residual-based architecture and hence the starting point of our work is the \textit{bottleneck} block described in \cite{he2016deep,he2016identity}. More recently, \cite{xie2016aggregated} explores the idea of increasing the cardinality of the residual block by splitting it into a series of $c$ parallel (and much smaller so that the number of parameters remains roughly the same) sub-blocks with the same topology which behave as an ensemble. Beyond bottleneck layers, Szegedy et. al. \cite{szegedy2015going} propose the inception block which introduces parallel paths with different receptive field sizes and various ways of lowering the number of parameters by factorizing convolutional layers with large filters into smaller ones. In a follow-up paper \cite{szegedy2017inception}, the authors introduce a number of inception-residual architectures. The latter work is the most related one to the proposed method.

Our method is different from the aforementioned architectures in the following ways (see Fig.~\ref{fig:layers_wider_intro}): we create a hierarchical, parallel and multi-scale structure that (a) increases the receptive field size inside the block and (b) improves gradient flow, (c) is specifically designed to have (almost) the same number of parameters as the original bottleneck, (d) our block does not contain $1 \times 1$ convolutions, and (e) our block is derived from the perspective of improving the performance and efficiency of binary networks.

\subsection{Network design} Our target was not to propose a new network architecture for landmark localization; hence we used the state-of-the-art \textit{Hour-Glass} (HG) network of \cite{newell2016stacked} which makes use of the bottleneck block of \cite{he2016deep}. Because we are interested in efficiency, most of our experiments are conducted using a single network. Our baseline was the single binary HG obtained by directly quantizing it using \cite{rastegari2016xnor}. As Table~\ref{tab:bootlneck_vs} shows, there is a significant performance gap between the binary and the real valued HGs. We bridge this gap by replacing the bottleneck block used in the original HG with the proposed block.

\subsection{Human Pose Estimation} Traditionally, human pose estimation methods relied on tree structured graphical models~\cite{eichner20122d,buehler2011upper,yang2011articulated,pishchulin2013strong,sapp2013modec,belagiannis20143d} to represent the spatial relationships between body parts and were usually built using hand crafted features. More recently, methods based on CNNs have shown remarkable results outperforming traditional methods by large margin~\cite{toshev2014deeppose,tompson2014joint,pfister2015flowing,insafutdinov2016deepercut,bulat2016human,newell2016stacked,wei2016convolutional,belagiannis2017recurrent}. Because learning a direct mapping from the image to the location of the body parts is a highly non-linear problem that is difficult to learn, most methods represent each landmark as a confidence map encoded as a 2D Gaussian centered at the landmark's location and adopt the fully convolutional framework of~\cite{long2015fully}. {\color{black}Furthermore, instead of making single-shot predictions, almost all methods follow a cascaded approach making a number of intermediate predictions, refined in a sequential manner~\cite{bulat2016human,newell2016stacked,wei2016convolutional}. Notably, to further reduce the number of parameters of the cascaded approaches the method introduced  in~\cite{belagiannis2017recurrent} uses a recurrent neural network.}

While achieving remarkable performance, all the aforementioned deep learning methods are computationally demanding, requiring at least one high-end GPU. In contrast, our network uses binary weights and activations and as such it is intended to run on systems with limited resources (e.g. embedded devices, smartphones).

\subsection{Face Alignment}
Current state-of-the-art for large pose 2D and 3D face alignment is also based on CNNs \cite{jourabloo2016large,bulat2016convolutional,bulat2016two,ranjan2017all,bulat2017far,wu2017godp}. However, despite their accuracy, these methods are computationally demanding. Our network produces state-of-the-art results for this task, yet it is designed to run on devices with limited computational resources.

\section{Background}\label{sec:method}

\begin{figure}[!htb]
    \centering
    \includegraphics[height=1.5in,trim={0.5cm 0.5cm 0.5cm 0.5cm},clip]{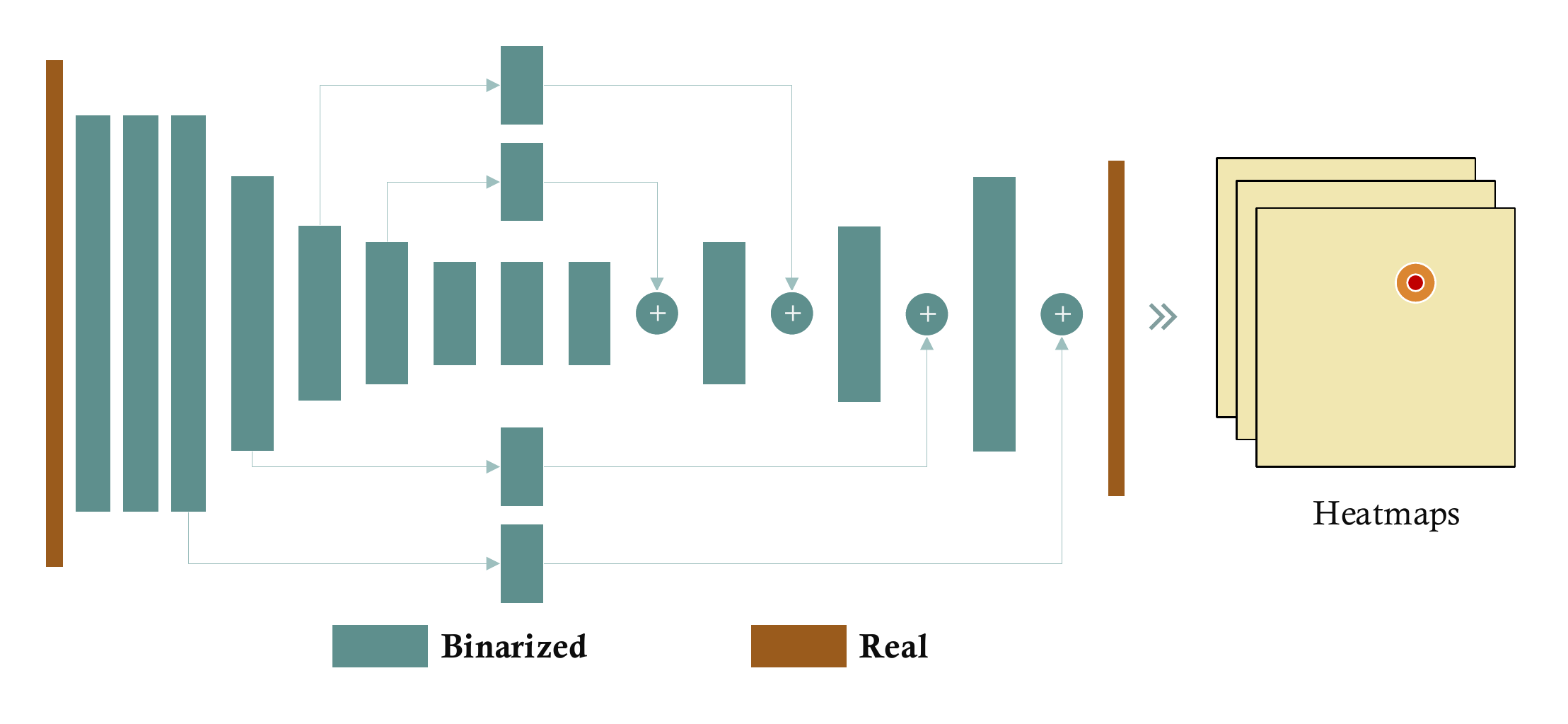}
    \caption{The architecture of a single \textit{Hour-Glass} (HG) network \cite{newell2016stacked}. Following \cite{rastegari2016xnor}, the first and last layers {\color{black}(brown colour)} are left real while all the remaining layers are binarized.}
    \label{fig:network_small}
\end{figure}

The ResNet consists of two types of blocks: \textit{basic} and \textit{bottleneck}. We are interested only in the latter one which was designed to reduce the number of parameters and keep the network memory footprint under control. We use the ``pre-activation'' version of \cite{he2016identity}, in which batch normalization \cite{ioffe2015batch} and the activation function precede the convolutional layer. Note that we used the version of bottleneck defined in \cite{newell2016stacked} the middle layer of which has 128 channels (vs 64 used in \cite{he2016identity}).

The residual block is the main building block of the Hourglass (HG) network, shown in Fig. \ref{fig:network_small}, which is a state-of-the-art architecture for landmark localization that predicts a set of heatmaps (one for each landmark) in a fully convolutional fashion. The HG network is an extension of \cite{long2015fully} allowing however for a more symmetric top-down and bottom-up processing. See also \cite{newell2016stacked}.

\section{Method}\label{S:Method}

Herein, we describe how we derive the proposed binary hierarchical, parallel and multi-scale block of Fig. \ref{fig:layers_parallel}. In Section~\ref{sec:binary}, by reducing the number of its parameters to match the ones of the original bottleneck, we further derive the block of Fig. \ref{fig:layers_wider_intro}. This Section is organized as follows:
\begin{itemize}
    \item
          We start by analyzing the performance of the binarized HG in Subsection \ref{ssec:naive} which provides the motivation as well as the baseline for our method.
    \item
          Then, we propose a series of architectural innovations in Subsections \ref{ssec:wide_block}, \ref{ssec:size_filters}, \ref{ssec:conv1} and \ref{ssec:better} (shown in Figs. \ref{fig:layers_wider}, \ref{fig:layers_mscale} and \ref{fig:layers_no1x1}) each of which is evaluated and compared against the binarized residual block of Subsection \ref{ssec:naive}.
    \item
          We continue, by combining ideas from these architectures, we propose the binary hierarchical, parallel and multi-scale block of Fig. \ref{fig:layers_parallel}. Note that the proposed block is not a trivial combination of the aforementioned architectures but a completely new structure.
    \item 
        {\color{black}Finally, we attempt to make a fair comparison between the performance of the proposed block against that of the original bottleneck module for both real and binary cases.}
\end{itemize}

We note that all results for this Section were generated for the task of human pose estimation   using the standard training-validation partition of MPII \cite{bulat2016human,newell2016stacked}.

\subsection{Binarized HG}\label{ssec:naive}

The binarization is accomplished using:
\begin{equation}
    \ma{I} * \ma{W} \approx (sign(\ma{I}) \circledast sign(\ma{W})) * \alpha \label{eq:additive0},
\end{equation}
where $\ma{I}$ is the input tensor, $\ma{W}$ represents the layer weights, {\color{black}$\alpha\in \mathbb{R}^{+}$ is a scaling factor computed as the average of the absolute weight values} and $\circledast$ denotes the binary convolution operation which can be efficiently implemented with XNOR.

We start from the original bottleneck blocks of the HG network and, following \cite{rastegari2016xnor}, we binarize them keeping only the first and last layers of the network real. See also Fig. \ref{fig:network_small}. This is crucial, especially for the very last layer where higher precision is required for producing a dense output (heatmaps). Note that these layers account for less than 0.01\% of the total number of parameters.

\begin{figure}[!htb]
    \centering
    \includegraphics[height=2.0in,trim={1.7cm 1.7cm 1.7cm 1.7cm},clip]{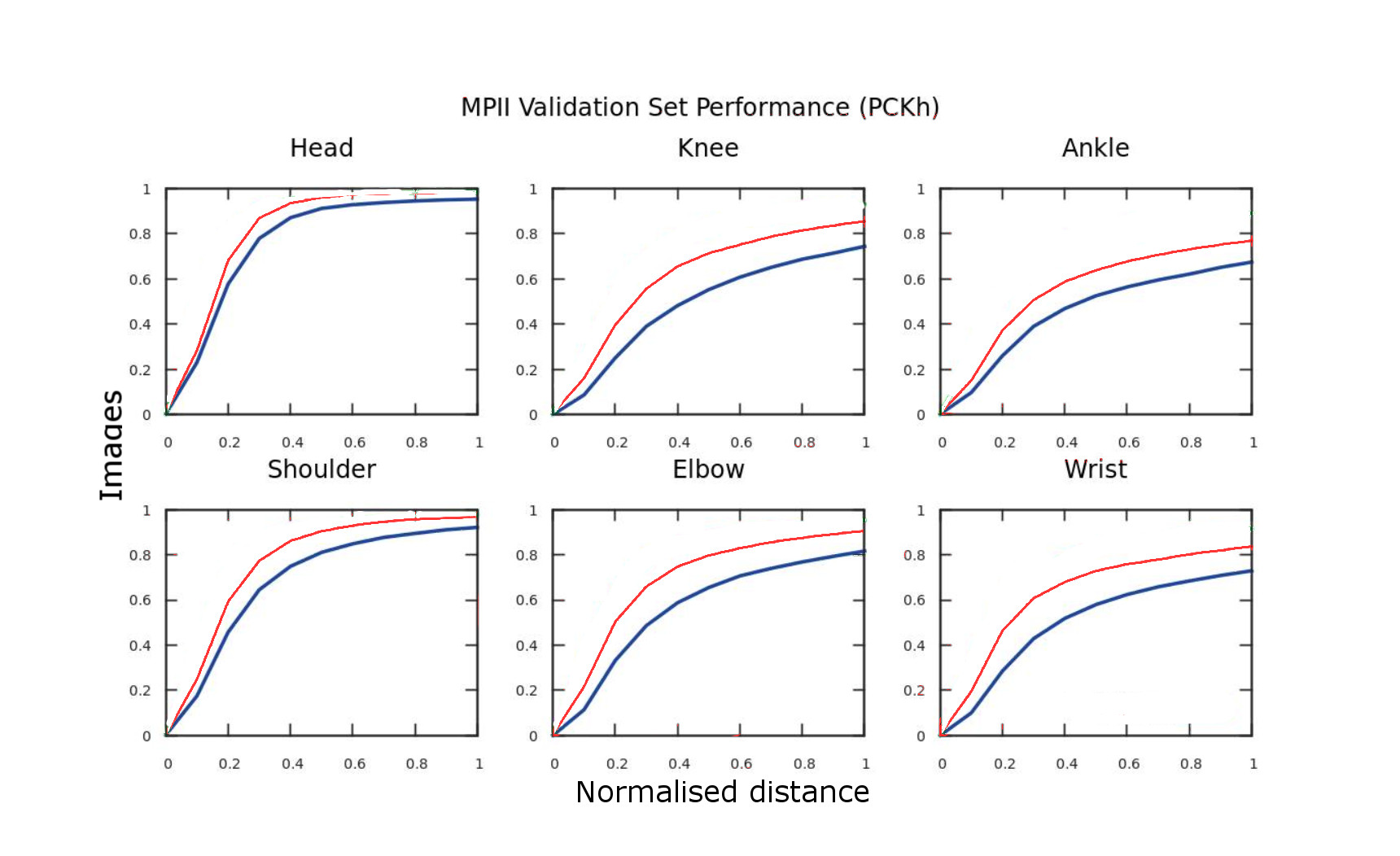}
    \caption{Cumulative error curves on MPII validation set for real-valued (red) and binary (blue) bottleneck blocks within the HG network.}
    \label{fig:naive_bech}
\end{figure}

\begin{figure}[!htb]
    \centering
    \includegraphics[height=1.5in,trim={0.5cm 0.5cm 0.5cm 0.5cm},clip]{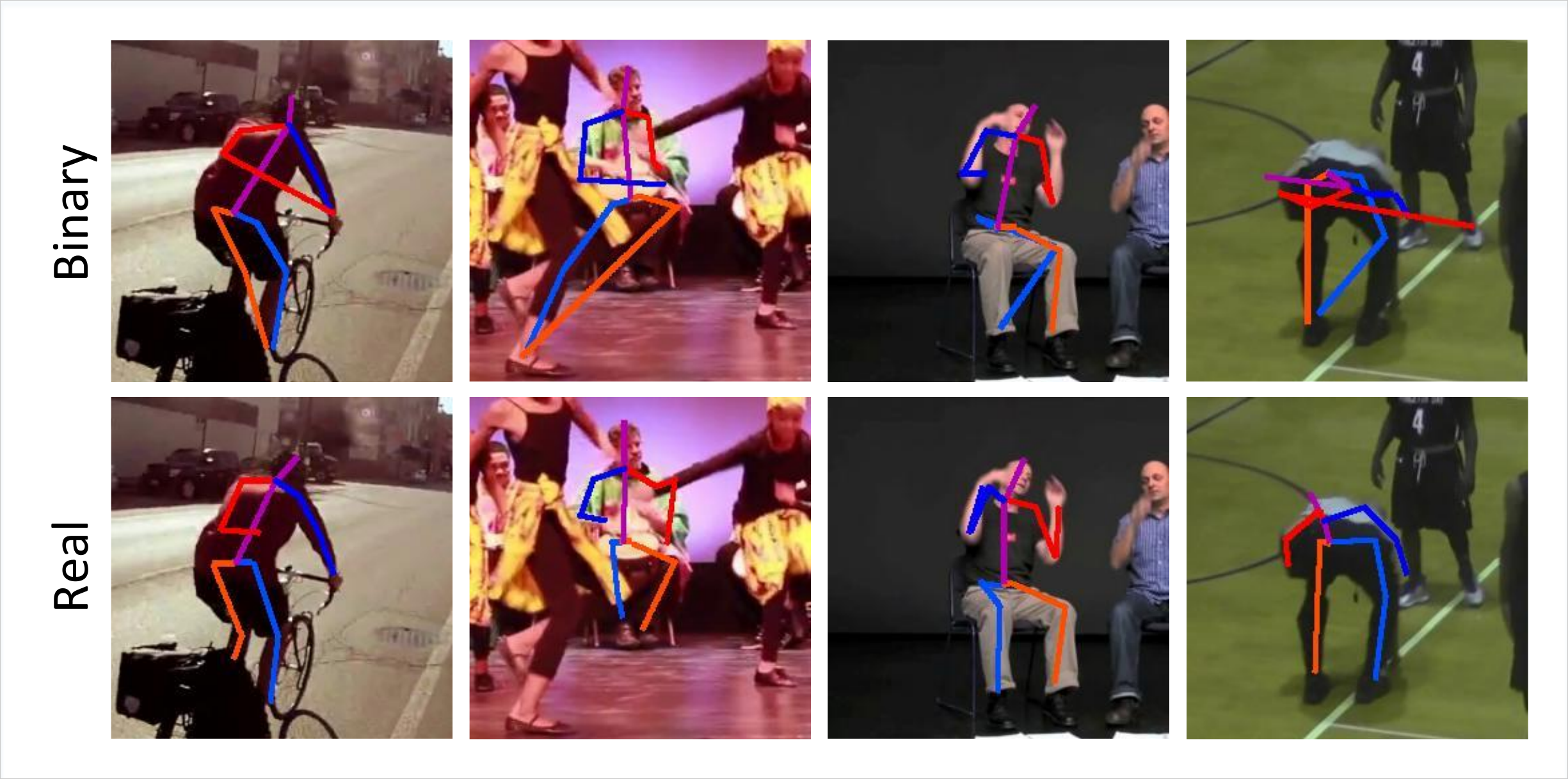}
    \caption{Examples of failure cases for the binarized HG (first row) and  predictions of its real-valued counterpart (second row). The binary HG misses certain range of poses while having similar accuracy for the correct parts.}
    \label{fig:fails}
\end{figure}

The performance of the original (real-valued) and the binarized HG networks can be seen in Fig. \ref{fig:naive_bech} and Table \ref{tab:bootlneck_vs}. We observe that binarization results in significant performance drop. As we may notice, for almost all parts, there is a large difference in performance which clearly indicates that the binary network has significant less representational power. Some failure cases are shown in Fig.~\ref{fig:fails} illustrating that the binary network was not able to learn some difficult poses. We address this with a better architecture as detailed in the next four Subsections.

\begin{table}[!htbp]
    \renewcommand{\arraystretch}{1.3}
    \caption{PCKh error on MPII dataset for real-valued and binary  bottleneck blocks within the HG network.}
    \label{tab:bootlneck_vs}
    \centering
    \begin{tabular}{|l|c|c|c|c|c|c|c|} 
        \hline
        Crit.   & Bottleneck (real) & Bottleneck (binary) \\
        \hline\hline
        Head    & 94.9              & 90.5                \\
        Shld    & 85.8              & 79.6                \\
        Elbow   & 76.9              & 63.0                \\
        Wrist   & 71.3              & 57.2                \\
        Hip     & 78.1              & 71.1                \\
        Knee    & 70.1              & 58.2                \\
        Ankle   & 63.2              & 53.4                \\
        \hline
        PCKh    & 76.5              & 67.2                \\
        \hline
        \# par. & 3.5M              & 3.5M                \\
        \hline
    \end{tabular}
\end{table}

\subsection{On the Width of Residual Blocks} \label{ssec:wide_block}

The original bottleneck block of Fig. \ref{fig:layers_original} is composed of 3 convolutional layers with a filter size of $1\times1$, $3\times3$ and $1\times1$, with the first layer having the role of limiting the width (i.e. the number of channels) of the second layer, thus greatly reducing the number of parameters inside the module. However, it is unclear whether the idea of having a bottleneck structure will be also successful for the binary case, too. Due to the limited representational power of the binary layers, greatly reducing the number of channels  might reduce the amount of information that can be passed from one layer to another, leading to lower performance.

To investigate this, we modify the bottleneck block by increasing the number of channels in the \textit{thin} $3\times3$ layer from 128 to 256. By doing so, we match the number of channels from the first and last layer, effectively removing the ``bottleneck'', and increasing the amount of information that can be passed from one block to another. The resulting \textbf{wider} block is shown in Fig.~\ref{fig:layers_wider}. Here, ``wider''\footnote {The term wider here strictly refers to a ``moderate'' increase in the number of channels in the \textit{thin} layer (up to 256), effectively removing the ``bottleneck''. Except for the naming there is no other resemblance with \cite{zagoruyko2016wide} which performs a study of wide vs deep, using a different building block alongside a much higher number of channels (up to 2048) and without any form of quantization. A similar study falls outside the scope of our work.}  refers to the increased number of channels over the initial \textit{thin} layer.

As Table \ref{tab:method_results} illustrates, while this improves performance against the baseline, it also raises the memory requirements. \newline \textbf{Conclusion}: Widening the \textit{thin} layer offers tangible performance improvement, however at a high computational cost.

\subsection{On Multi-Scale Filtering} \label{ssec:size_filters}
\begin{figure}[!htb]
    \centering
    \includegraphics[height=0.6in,trim={0.5cm 0.5cm 0.5cm 0.5cm},clip]{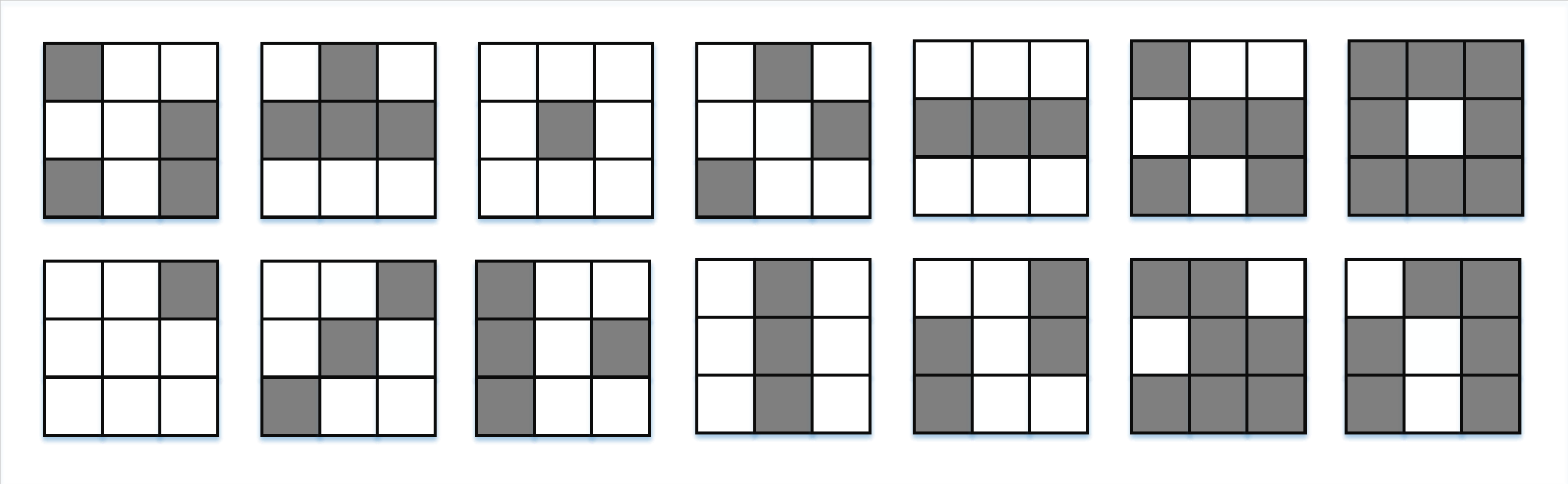}
    \caption{Examples of learned $3 \times 3$ binary filters.}
    \label{fig:ifilters_3x3}
\end{figure}

Small filters have been shown both effective and efficient \cite{simonyan2014very,szegedy2015going} with models being solely made up by a combination of convolutional layers with $3 \times 3$ and/or $1 \times 1$ filters \cite{he2016deep,he2016identity,simonyan2014very}. For the case of real-valued networks, a large number of kernels can be learned. However, for the binary case, the number of possible unique convolutional kernels is limited to $2^k$ states only, where $k$ is the size of the filter.
Examples of such $3 \times 3$ learned filters are shown in Fig.~\ref{fig:ifilters_3x3}.

To address the limited representation power of $3 \times 3$ filters for the binary case, and similarly to \cite{szegedy2017inception}, we largely depart from the block of Fig.~\ref{fig:layers_wider} by proposing the multi-scale structure of Fig.~\ref{fig:layers_mscale}. Note that we implement our multi-scale approach using both larger filter sizes and max-pooling, which greatly increase the effective receptive field within the block. Also, because our goal is to analyze the impact of a multi-scale approach alone, we intentionally keep the number of parameters to a similar level to that of the original bottleneck block of  Fig.~\ref{fig:layers_original}. To this end, we avoid a leap in the number of parameters, by (a) decomposing the $5 \times 5$ filters into two layers of $3 \times 3$  filters, and (b) by preserving the presence of \textit{thin} layer(s) in the middle of the block.

Given the above, we split the input into two branches. The first (left) branch works at the same scale as the original bottleneck of Fig.~\ref{fig:layers_original} but has a $1 \times 1$ layer that projects the 256 channels into 64 (instead of 128) before going to the $3\times3$ one. The second (right) branch performs a multi-scale analysis by firstly passing the input through a max-pooling layer and then creating two branches, one using a $3 \times 3$ filter and a second one using a $5 \times 5$ decomposed into two $3 \times 3$. By concatenating the outputs of these two sub-branches, we obtain the remaining 64 channels (out of the 128 of the original bottleneck block). Finally, the two main branches are concatenated adding up to 128 channels, which are again back-projected to 256 with the help of a convolutional layer with $1 \times 1$ filters.

The accuracy of the proposed structure can be found in Table~\ref{tab:method_results}. We can observe a healthy performance improvement at little additional cost and similar computational requirements to the original bottleneck of Fig.~\ref{fig:layers_original}. \newline\textbf{Conclusion}: When designing binarized networks, multi-scale filters should be preferred.

\subsection{On $1 \times 1$ Convolutions}\label{ssec:conv1}

In the previously proposed block of Fig.~\ref{fig:layers_mscale}, we opted to avoid an increase in the number of parameters, by retaining the two convolutional layers with $1 \times 1$ filters. In this Subsection, by relaxing this restriction, we analyze the influence of $1 \times 1$ filters on the overall network performance.

\begin{figure}[!htb]
    \centering
    \includegraphics[height=1.2in,trim={0.1cm 0.1cm 0.1cm 0.1cm},clip]{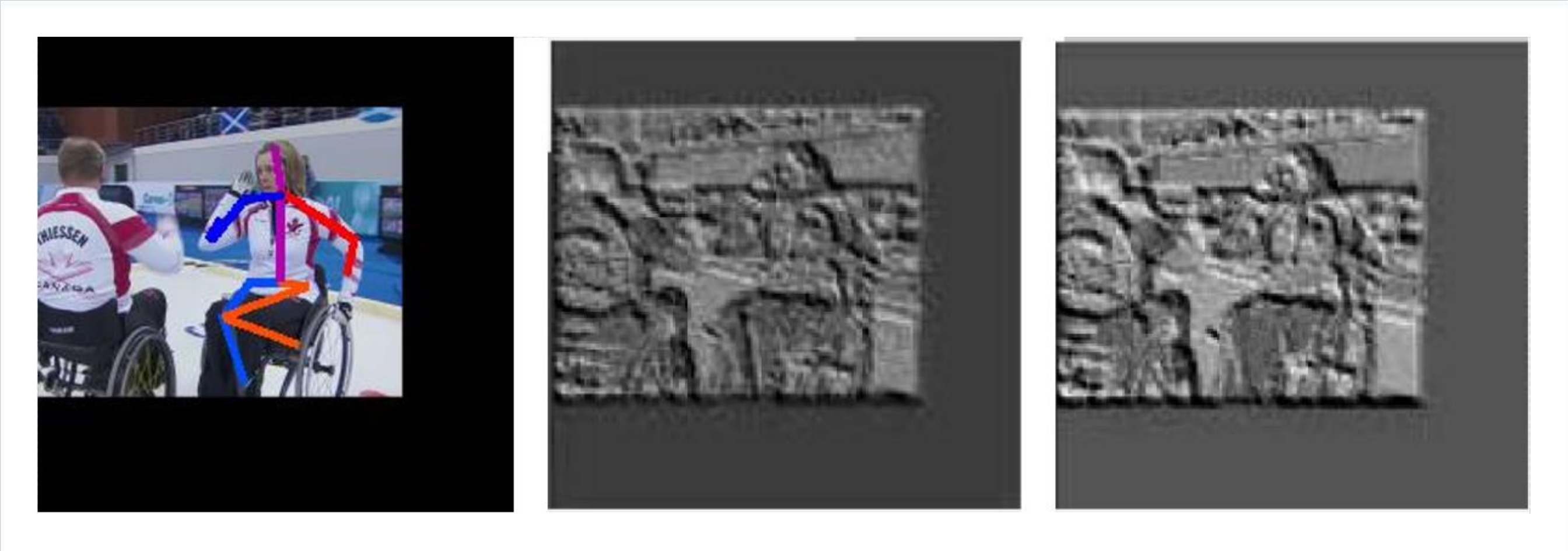}
    \caption{Examples of features before and after a $1 \times 1$ convolutional layer. Often the features are copied over with little modifications, usually consisting in the details' removal. The contrast was altered for better visualization.}
    \label{fig:ifilters_1x1}
\end{figure}

In particular, we remove all convolutional layers with $1 \times 1$ filters from the multi-scale block of Fig.~\ref{fig:layers_mscale}, leading to the structure of  Fig.~\ref{fig:layers_no1x1}. Our motivation to remove $1 \times 1$ convolutions for the binary case is the following: because $1 \times 1$ filters are limited to two states only (either 1 or -1) they have a very limited learning power. Due to their nature, they behave as simple filters deciding when a certain value should be passed or not. In practice, this allows the input to pass through the layer with little modifications, sometimes actually blocking ``good features'' and hurting the overall performance by a noticeable amount. This is particularly problematic for the task of landmark localization, where a high level of detail is required for successful localization. Examples of this problem are shown in Fig.~\ref{fig:ifilters_1x1}.

Results reported in Table~\ref{tab:method_results} show that by removing $1 \times 1$ convolutions, performance over the baseline is increased by more than 8\%. Even more interestingly, the newly introduced block outperforms the one of Subsection~\ref{ssec:wide_block}, while having less parameters, which shows that the presence of $1 \times 1$ filters limits the performance of binarized CNNs. \newline \textbf{Conclusion}: The use of $1 \times 1$ convolutional filters on binarized CNNs has a detrimental effect on performance and should be avoided.

\subsection{On Hierarchical, Parallel \& Multi-Scale}\label{ssec:better}

\begin{figure*}[!htb]
    \centering
    \subfloat[The \textbf{Original Bottleneck} block with pre-activation, as defined in \cite{he2016identity}. Its binarized version is described in Section~\ref{ssec:naive}.]{\makebox[0.3\textwidth]{\includegraphics[height=2in,trim={0.5cm 0.5cm 0.5cm 0.5cm},clip]{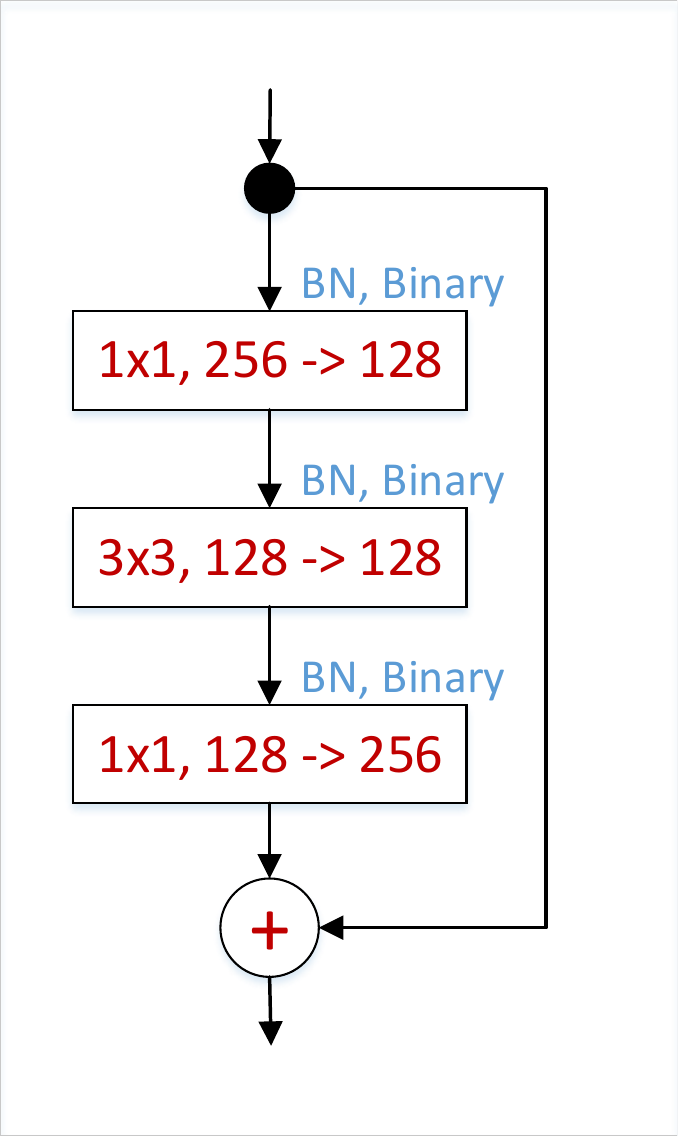}}
        \label{fig:layers_original}}
    \hfill
    \subfloat[The \textbf{Wider} version of (a) produced by increasing the number of filters in the second layer. See Subsection~\ref{ssec:wide_block}.]{\makebox[0.3\textwidth]{\includegraphics[height=2in,trim={0.5cm 0.5cm 0.5cm 0.5cm},clip]{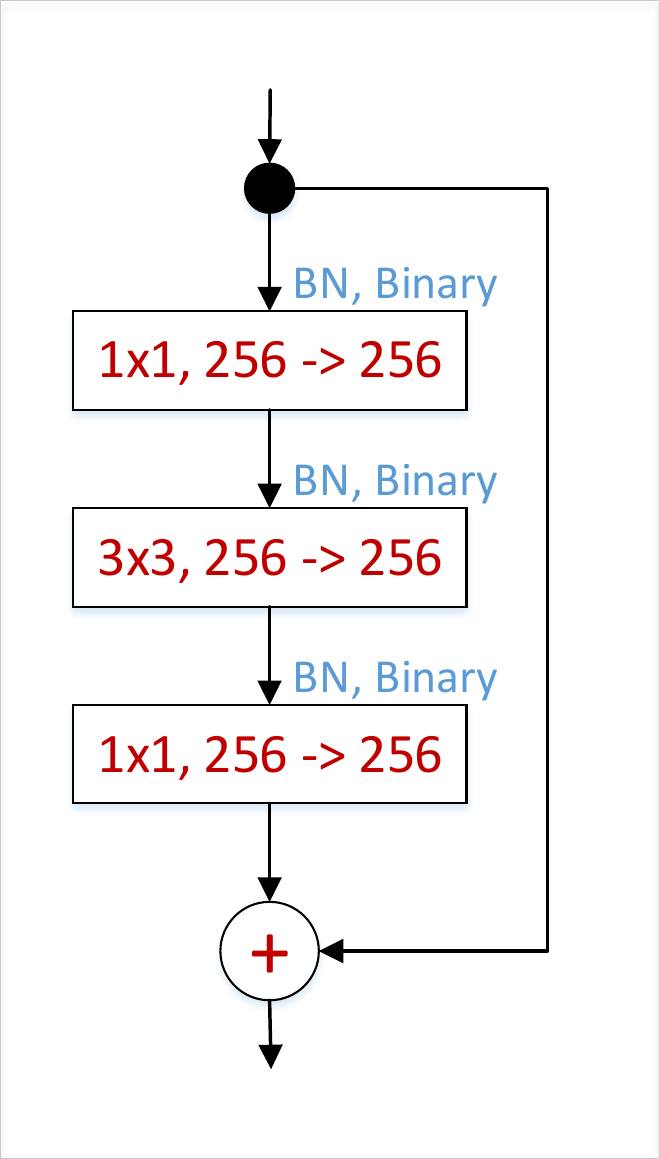}}
        \label{fig:layers_wider}}
    \hfill
    \subfloat[Largely departing from (b), this block consists of \textbf{Multi-Scale (MS)} filters for analyzing the input at multiple scales. See Subsection~\ref{ssec:size_filters}.]{\makebox[0.3\textwidth]{\includegraphics[height=2in,trim={0.5cm 0.5cm 0.5cm 0.5cm},clip]{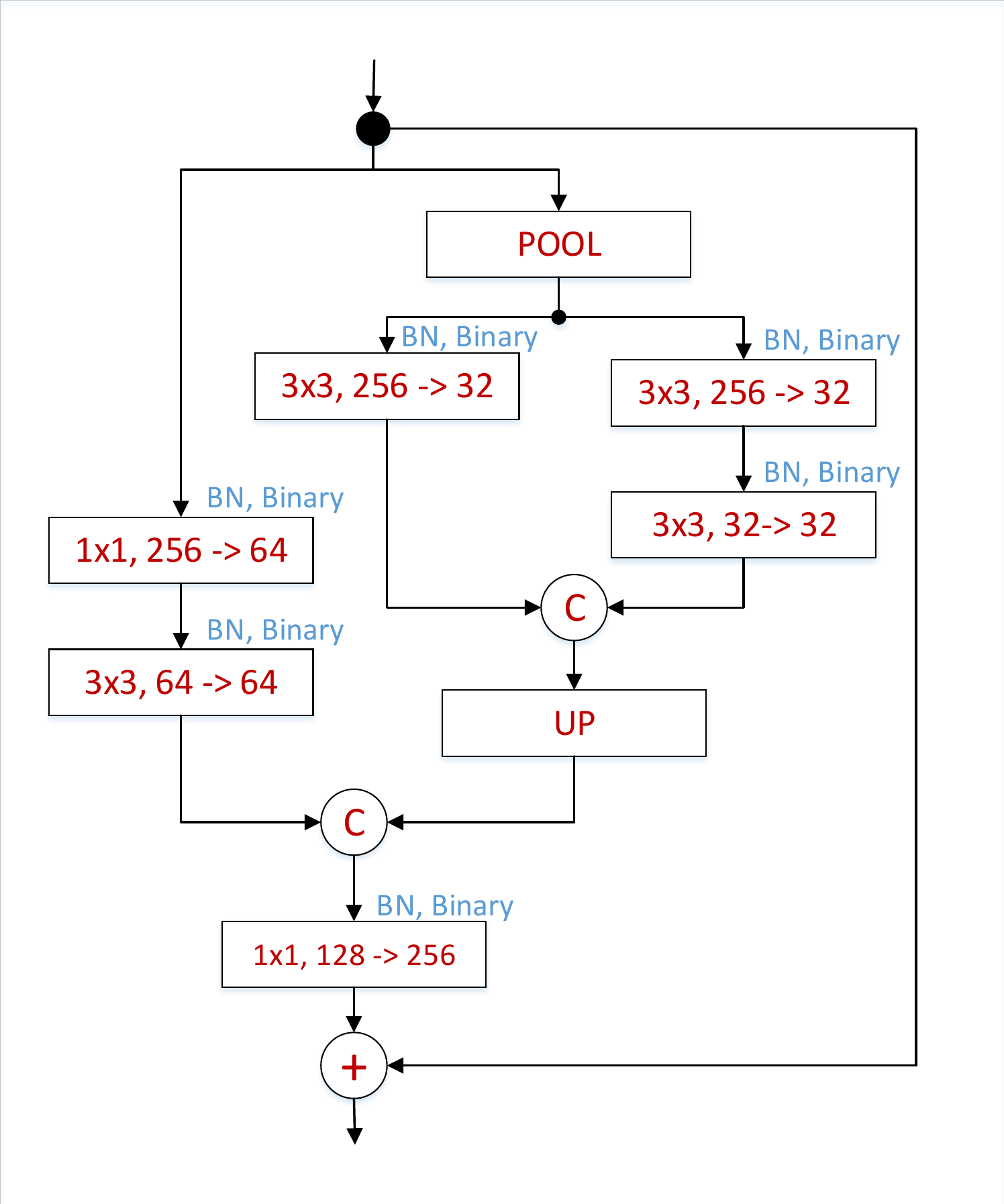}}
        \label{fig:layers_mscale}}

    \bigskip

    \subfloat[A variant of the MS block introduced in (c) after removing all convolutional layers with $1\times 1$ filters (\textbf{MS Without $1\times 1$ filters}). See  Subsection~\ref{ssec:size_filters}.]{\makebox[0.45\textwidth]{\includegraphics[height=2in,trim={0.5cm 0.5cm 0.5cm 0.5cm},clip]{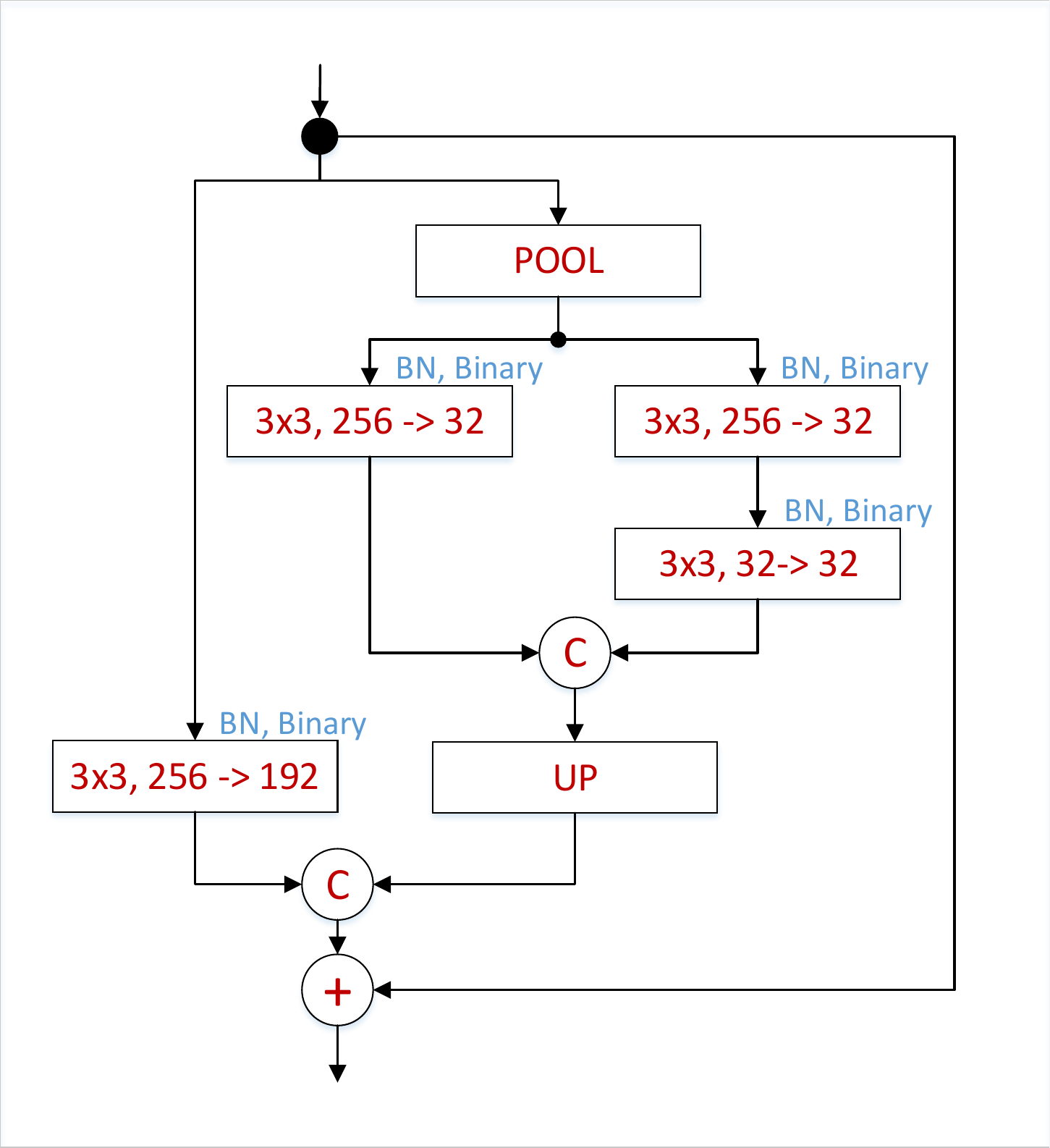}}
        \label{fig:layers_no1x1}}
    \hfill
    \subfloat[The proposed \textbf{Hierarchical, Parallel \& MS} (denoted in the paper as (\textbf{Ours, final}) block incorporates all ideas from (b), (c) and (d) with an improved gradient flow. See  Subsection~\ref{ssec:better}]{\makebox[0.45\textwidth]{\includegraphics[height=2in,trim={0.5cm 0.5cm 0.5cm 0.5cm},clip]{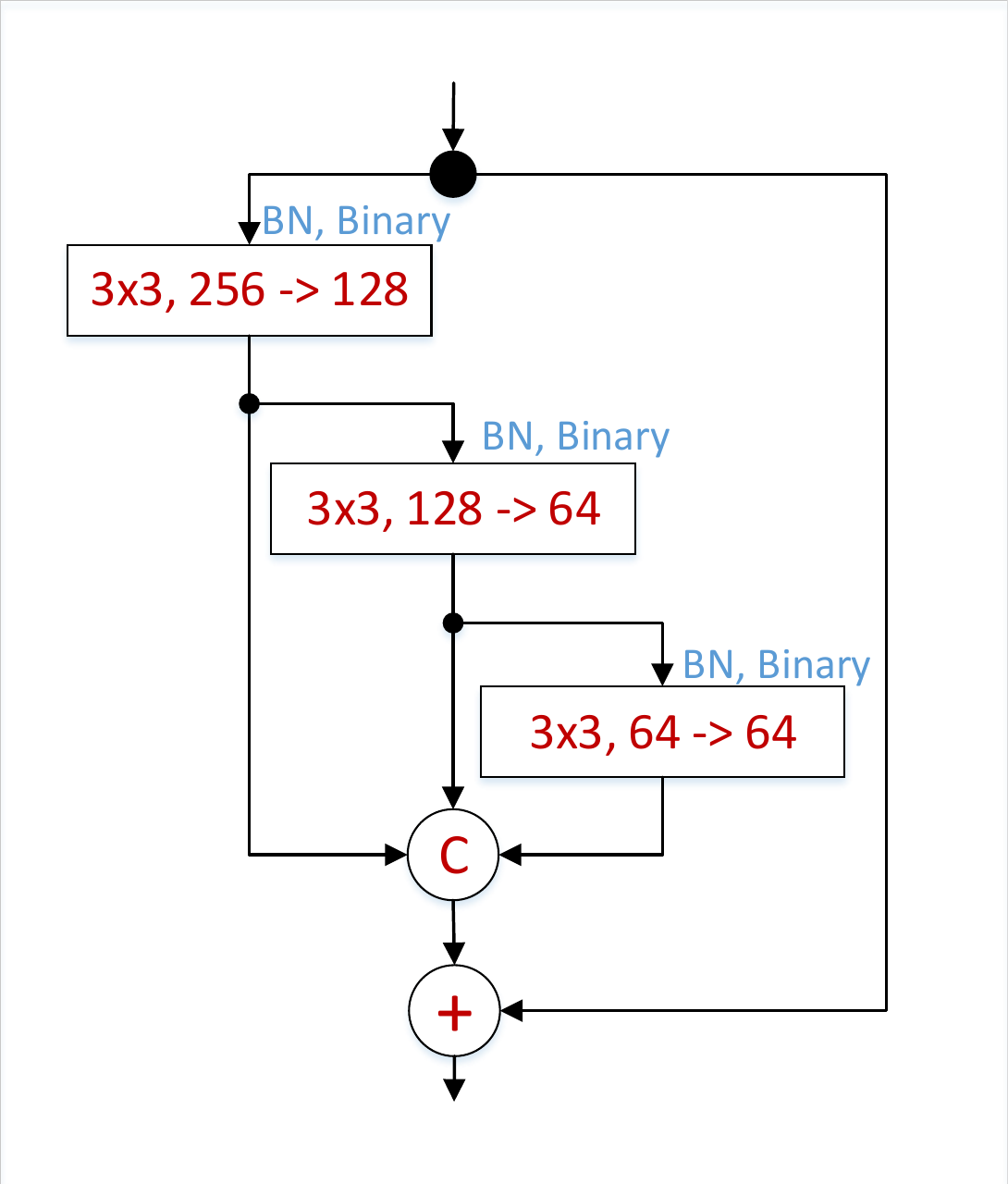}}
        \label{fig:layers_parallel}}

    \caption{Different types of blocks described and evaluated. Our best performing block is shown in figure (e). A layer is depicted as a rectangular block containing: its filter size, number of input channels and the number of output channels). ``C" - denotes concatenation operation, ``+" an element-wise sum {\color{black}and ``UP" a bilinearly upsample layer.}}
    \label{fig:layers}
\end{figure*}

Binary networks are even more sensitive to the problem of fading gradients \cite{courbariaux2016binarized,rastegari2016xnor}, and for our network we found that the gradients are up to 10 times smaller than those corresponding to its real-valued counterpart. To alleviate this, we design a new module which has the form of a hierarchical, parallel multi-scale structure allowing, for each resolution, the gradients to have 2 different paths to follow, the shortest of them being always 1. The proposed block is depicted in Fig.~\ref{fig:layers_parallel}. Note that, in addition to better gradient flow, our design encompasses all the findings from the previous Subsections: (a) no convolutional layers with $1 \times 1$ filters should be used, (b) the block should preserve its width as much as possible (avoiding large drops in the number of channels), and (c) multi-scale filters should be used.

Contrary to the blocks described in Subsections~\ref{ssec:wide_block} -~\ref{ssec:conv1}, where the gradients may need to pass through two more layers before reaching the output of the block, in the newly proposed module, each convolutional layer has a direct path that links it to the output, so that at any given time and for all the layers within the module the shortest possible path is equal to 1. The presence of a hierarchical structure inside the module efficiently accommodates larger filters (up to $7 \times 7$), decomposed into convolutional layers with $3 \times 3$ filters. {\color{black} This allows for the information to be analysed at different scales because of the different filter sizes used  (hence the term ``multi-scale''). We opted not to use pooling because it results in loss of information. } Furthermore, our design avoids the use of an element-wise summation layer as for example in \cite{xie2016aggregated,szegedy2017inception}, further improving the gradient flow and keeping the complexity under control.

As we can see in Table~\ref{tab:method_results}, the proposed block matches and even outperforms the block proposed in Section~\ref{ssec:size_filters} having far less parameters.
\begin{table}[!htbp]
    \renewcommand{\arraystretch}{1.3}
    \caption{PCKh-based comparison of different blocks on MPII validation set. \# params refers to the number of parameters of the whole network.}
    \label{tab:method_results}
    \centering
    \begin{tabular}{|l|c|c|}
        \hline
        Block type                                             & \# params     & PCKh          \\
        \hline\hline
        Bottleneck (original) (Fig.~\ref{fig:layers_original}) & 3.5M          & 67.2\%        \\
        \hhline{|=|=|=|}
        Wider (Fig.~\ref{fig:layers_wider})                    & 11.3M         & 70.7\%        \\
        Multi-Scale (MS) (Fig.~\ref{fig:layers_mscale})        & 4.0M          & 69.3\%        \\
        MS without 1x1 filters (Fig.~\ref{fig:layers_no1x1})   & 9.3M          & 75.5\%        \\
        Bottleneck (wider) + no $1 \times 1$                    & 5.8M          & 69.5\% 
            \\
        \hhline{|=|=|=|}
        {\begin{tabular}{c}
                \textbf{Hierarchical, Parallel \& MS } \\ \textbf{(Ours, Final)} (Fig.~\ref{fig:layers_wider_intro})\end{tabular}}

                                                               & \textbf{4.0M} & \textbf{72.7\%} \\
        \hline
        {\begin{tabular}{c}
                \textbf{Hierarchical, Parallel \& MS } \\ \textbf{(Ours, Final)} (Fig.~\ref{fig:layers_parallel})\end{tabular}}

                                                               & \textbf{6.2M} & \textbf{76\%} \\
        \hline
    \end{tabular}
\end{table}
\newline \textbf{Conclusion}: Good gradient flow and hierarchical multi-scale filtering are crucial for high performance without excessive increase in the parameters of the binarized network.

\subsection{Proposed vs Bottleneck} \label{sec:binaryvsreal}
In this Section, we attempt to make a fair comparison between the performance of the proposed block (\textbf{Ours, Final}, as in Fig.~\ref{fig:layers_parallel}) against that of the original bottleneck module (Fig.~\ref{fig:layers_original}) by taking two important factors into account:
\begin{itemize}
    \item
          Both blocks should have the same number of parameters.
    \item
          The two blocks should be compared for the case of binary but also real-valued networks.
\end{itemize}
With this in mind, in the following Sections, we show that:
\begin{itemize}
    \item
          The proposed block largely outperforms a bottleneck with the same number of parameters for the binary case.
    \item
          The proposed block also outperforms a bottleneck with the same number of parameters for the real case but in this case the performance difference is smaller.
\end{itemize}
We conclude that, for the real case, increasing the number of parameters (by increasing width) results in performance increase; however this is not the case for binary networks where a tailored design as the one proposed here is needed.

\subsubsection{Binary} \label{sec:binary}

To match the number of parameters between the proposed and bottleneck block, we follow two paths. Firstly, we increase the number of parameters of the bottleneck: (a) a first way to do this is to make the block wider as described in Section \ref{ssec:wide_block}. Note that in order to keep the number or input-output channels equal to 256, the resulting block of Fig. \ref{fig:layers_wider} has a far higher number of parameters than the proposed block. Despite this, the performance gain is only moderate (see Section \ref{ssec:wide_block} and Table~\ref{tab:method_results}). (b) Because we found that the $1 \times 1$ convolutional layers have detrimental effect to the performance of the Multi-Scale block of Fig.~\ref{fig:layers_mscale}, we opted to remove them from the bottleneck block, too. To this end, we modified the Wider module by (a) removing the $1 \times 1$ convolutions and (b) halving the number of parameters in order to match the number of parameters of the proposed  block. The results in Table~\ref{tab:method_results} clearly show that this modification is helpful but far from being close to the performance achieved by the proposed block.

Secondly, we decrease the number of parameters in the proposed block to match the number of parameters of the original bottleneck. This block is shown in Fig. \ref{fig:layers_wider_intro}. To this end, we reduced the number of input-output channels of the proposed block from 256 to 192 so that the number of channels in the first layer are modified from [256 $\rightarrow$ 128, $3\times3$] to [192$\rightarrow$96, $3\times3$], in the second layer from [128$\rightarrow$64, $3\times3$] to [96$\rightarrow$48, $3\times3$] and in the third layer from [64$\rightarrow$64, $3\times3$] to [48$\rightarrow$48, $3\times3$].  Notice, that even in this case, the proposed binarized module outperforms the original bottleneck block by more than 5\% (in absolute terms) while both have very similar number of parameters (see Table~\ref{tab:method_results}). 	

\subsubsection{Real} \label{sec:real}

While the proposed block was derived from a binary perspective, Table~\ref{tab:real} shows that a significant performance gain is also observed for the case of real-valued networks. In order to quantify this performance improvement and to allow for a fair comparison, we increase the number of channels inside the original bottleneck block so that both networks have the same depth and a similar number of parameters. {\color{black} For our binary block, in order to bring it back to the real valued domain, we simply replace the ``sign'' function with ReLU activations while keeping all the weights real.} Even in this case, our block outperforms the original block although the gain is smaller than that observed for the binary case. We conclude that for real-valued networks performance increase can be more easily obtained by simply increasing the number of parameters, but for the binary case a better design is needed as proposed in this work.

\begin{table}[!htbp]
    \renewcommand{\arraystretch}{1.3}
    \caption{PCKh-based performance on MPII validation set for real-valued blocks: Our block is compared with a wider version of the original bottleneck so that both blocks have similar \# parameters.}
    \label{tab:real}
    \centering
    \begin{tabular}{|l|c|c|}
        \hline
        Layer type             & \# parameters & PCKh            \\
        \hline\hline
        Bottleneck (wider)     & 7.0M          & 83.1\%          \\
        \hline
        \textbf{(Ours, Final)} & \textbf{6.2M} & \textbf{85.5\%} \\
        \hline
    \end{tabular}
\end{table}

\section{Ablation studies} \label{sec:ablation}

In this Section, we present a series of other architectural variations and their effect on the performance of our binary network. All reported results are obtained using the proposed block of Fig.~\ref{fig:layers_parallel} coined \textbf{Ours, Final}. We focus on the effect of augmentation and different losses which are novel experiments not reported in \cite{rastegari2016xnor}, and then comment on the effect of pooling, ReLUs and performance speed-up. 

\textbf{Is Augmentation required?} Recent works have suggested that binarization is an extreme case of regularization \cite{courbariaux2015binaryconnect,courbariaux2016binarized,merolla2016deep}. In light of this, one might wonder whether data augmentation is still required.  Table~\ref{tab:aug_results} shows that in order to accommodate the presence of new poses and/or scale variations, data augmentation is very helpful providing a large increase (4\%) in performance. {\color{black} See Section \ref{sec:training} for more details on how augmentation was performed}.

\begin{table}[!htbp]
    \renewcommand{\arraystretch}{1.3}
    \caption{The effect of using:  augmentation, different losses (Sigmoid vs L2), different pooling methods and of adding a ReLU after the conv layer,  when training our binary network in terms of PCKh-based performance on MPII validation set. {\color{black}We note that ``(Ours, Final)" was trained using a Sigmoid Loss, Maxpooling and applying augmentation. The additional text after it denotes the change made.}}

    \label{tab:aug_results}
    \centering
    \begin{tabular}{|l|c|c|}
        \hline
        Layer type                    & \# parameters & PCKh          \\
        \hline\hline
        (Ours, Final) - No Aug.       & 6.2M          & 72.1\%        \\
        \hline
        (Ours, Final) - L2 loss       & 6.2M          & 73.8\%        \\
        \hline
        (Ours, Final) - AvgPool      & 6.2M          & 71.9\%        \\
        \hhline{|=|=|=|}
        
        \textbf{(Ours, Final)} & \textbf{6.2M}  & \textbf{76\%} \\
        
        \hline
        \textbf{(Ours, Final) + ReLU} & \textbf{6.2M} & \textbf{78.1\%} \\
        \hline
    \end{tabular}

\end{table}

\textbf{The effect of loss}. We trained our binary network to predict a set of heatmaps, one for each landmark \cite{tompson2014joint}. To this end, we experimented with two types of losses: the first one places a Gaussian around the correct location of each landmark and trains using a pixel-wise L2 loss \cite{tompson2014joint}. However, the gradients generated by this loss are usually small even for the case of a real-valued network. Because binarized networks tend to amplify this problem, as an alternative, we also experimented with the Sigmoid cross-entropy pixel-wise loss typically used for detection tasks \cite{zhang2015fine}. We found that the use of the Sigmoid cross-entropy pixel-wise loss increased the gradients by 10-15x (when compared to the L2 loss), offering a 2\% improvement (see Table~\ref{tab:aug_results}), after being trained for the same number of epochs.
 
\textbf{Pooling type.} In the context of binary networks, and because the output is restricted to 1 and -1, max-pooling might result in outputs full of 1s only. To limit this effect, we placed the activation function before the convolutional layers as proposed in \cite{he2016identity,rastegari2016xnor}. Additionally, we opted to replace max-pooling with average pooling. However, this leads to slightly worse results (see Table~\ref{tab:aug_results}). In practice, we found that the use of blocks with pre-activation suffices and that the ratio of 1 and -1 is close to 50\% even after max-pooling.

\textbf{With or without ReLU.} Because during the binarization process all ReLU layers are replaced with the Sign function, one might wonder if ReLUs are still useful for the binary case. Our findings are in line with the ones reported in \cite{rastegari2016xnor}. By adding a ReLU activation after each convolutional layer, we observe a 2\% performance improvement (see Table~\ref{tab:aug_results}), which can be attributed to the added non-linearity, particularly useful for training very deep architectures.

\textbf{Performance.} In theory, by replacing all floating-point multiplications with bitwise XOR and making use of the SWAR (Single instruction, multiple data within a register) \cite{rastegari2016xnor,courbariaux2016binarized}, the number of operations can be reduced up to 32x when compared against the multiplication-based convolution. However, in our tests, we observed speedups of up to 3.5x, when compared against cuBLAS, for matrix multiplications, a result being in accordance with those reported in \cite{courbariaux2016binarized}. We note that we did not conduct experiments on CPUs. However, given the fact that we used the same method for binarization as in \cite{rastegari2016xnor}, similar improvements in terms of speed, of the order of 58x, are to be expected: as the real-valued network takes 0.67 seconds to do a forward pass on a i7-3820 using a single core, a speedup close to x58 will allow the system to run in real-time.

In terms of memory compression, by removing the biases, which have minimum impact (or no impact at all) on performance, and by grouping and storing every 32 weights in one variable, we can achieve a compression rate of 39x when compared against the single precision counterpart of Torch.

\section{Comparison with state-of-the-art} \label{sec:results}

In this Section, we compare our method against the current state-of-the-art for human pose estimation and 3D face alignment. Our final system comprises a single HG network but replaces the real-valued bottleneck block used in \cite{newell2016stacked} with the proposed binary, parallel, multi-scale block trained with the improvements detailed in Section~\ref{sec:ablation}. 

\subsection{Training} \label{sec:training}

All human pose estimation and 3D face alignment models were trained from scratch following the algorithm described in \cite{rastegari2016xnor} and using rmsprop \cite{tieleman2012lecture}. The initialization was done as in \cite{he2016deep}. For human pose estimation, we randomly augmented the data with rotation (between -$40^o$ and $40^o$ degrees), flipping and scale jittering (between 0.7 and 1.3). We  trained the network for 100 epochs, dropping the learning rate four times, from 2.5e-4 to 5e-5. A similar procedure was applied to the models for 3D face alignment, with the difference that the training was done for 55 epochs only. The input was normalized between 0 and 1 and all described networks were trained using the binary cross-entropy loss, defined as:

\begin{equation}
	\ve{l} = \frac{1}{N}\sum_{n=1}^{N}\sum_{i=1}^{W}\sum_{j=1}^{H}\ve{p}_{ij}^{n}\log{\ve{\widehat{p}}_{ij}^{n}} + (1 - \ve{p}_{ij}^{n})\log{(1 - \ve{\widehat{p}}_{ij}^{n})},
\end{equation}
where $\ve{p}_{ij}^{n}$ denotes the ground truth confidence map of the $n-$th part at the output pixel location $(i,j)$ and $\ve{\widehat{p}}_{ij}^{n}$ is the corresponding predicted output at the same location.

The models were implemented with Torch7 \cite{collobert2011torch7}.

\subsection{Human Pose Estimation.} As in all previous experiments, we used the standard training-validation partition of MPII \cite{bulat2016human,newell2016stacked}. We report the performance of (a) the proposed binary block, (b) the proposed block when implemented and trained with real values, (c) the real-valued stacked HG network consisting of 8 stacked single real-valued HG networks trained with intermediate supervision (state-of-the-art on MPII {\cite{newell2016stacked}}) and, finally, (d) the same real-valued network as in (c) where the bottleneck block is replaced by our proposed block.

The results are shown in Table~\ref{tab:mpii_results}. We observe that when a single HG network with the proposed block is trained with real weights, its performance reaches that of \cite{newell2016stacked}. This result clearly illustrates the enhanced learning capacity of the proposed block. Moreover, there is still a gap between the binary and real-valued version of the proposed block indicating that margin for further improvement is possible. We also observe that a full-sized model (with 8 HG networks) based on the proposed block performs slightly better than the original network from {\cite{newell2016stacked}}, indicating that, for the real-valued case, the new block is more effective than the original one when a smaller computational budget is used.

\begin{table}[!htbp]
    \renewcommand{\arraystretch}{1.3}
    \caption{PCKh-based comparison on MPII validation set. {\color{black}For ``Ours, bin.'' we report the results of its best variation, which includes the ReLU layer introduced in Section~\ref{sec:ablation}.}}
    \label{tab:mpii_results}
    \centering
    \begin{tabular}{|l|c|c|c|c|c|c|c|} 
        \hline
        Crit.   & \cite{newell2016stacked} & Ours, bin. & Ours[1x], real & Ours[8x], real \\
        \hline\hline
        Head    & 97.3                     & 94.7       & 96.8           & 97.4           \\
        Shld    & 96.0                     & 89.6       & 93.8           & 96.0           \\
        Elbow   & 90.2                     & 78.8       & 86.4           & 90.7           \\
        Wrist   & 85.2                     & 71.5       & 80.3           & 86.2           \\
        Hip     & 89.1                     & 79.1       & 87.0           & 89.6           \\
        Knee    & 85.1                     & 70.5       & 80.4           & 86.1           \\
        Ankle   & 82.0                     & 64.0       & 75.7           & 83.2           \\
        \hline
        PCKh    & 89.3                     & 78.1       & 85.5           & 89.8           \\
        \hline
        \# par. & 25M                      & 6M         & 6M             & 25M            \\
        \hline
    \end{tabular}
\end{table}

\subsection{Face alignment.} We used three very challenging datasets for large pose face alignment, namely AFLW {\cite{kostinger2011annotated}}, AFLW-PIFA {\cite{jourabloo2015pose}}, and AFLW2000-3D {\cite{zhu2016face}}. The evaluation metric is the Normalized Mean Error (NME) {\cite{jourabloo2015pose}.}

AFLW is a large-scale face alignment dataset consisting of 25,993 faces annotated with up to 21 landmarks. The images are captured in arbitrary conditions exhibiting a large variety of poses and expressions. As Table{~\ref{tab:aflw_full_results}} shows, our binarized network outperforms the state-of-the-art methods of \cite{ranjan2016hyperface} and \cite{ranjan2017all}, both of which use large real-valued CNNs.

\begin{table}[!htbp]
    \renewcommand{\arraystretch}{1.3}
    \caption{NME-based (\%) comparison on AFLW test set. The evaluation is done on the test set used in \cite{ranjan2017all}. }
    \label{tab:aflw_full_results}
    \centering
    \begin{tabular}{|l|c|c|c|c|} 
        \hline
        Method                               & [0,30]        & [30,60]       & [60,90]       & mean          \\
        \hline\hline
        HyperFace \cite{ranjan2016hyperface} & 3.93          & 4.14          & 4.71          & 4.26          \\
        \hline
        AIO \cite{ranjan2017all}             & 2.84          & 2.94          & 3.09          & 2.96          \\
        \hline
        \textbf{Ours}                        & \textbf{2.77} & \textbf{2.86} & \textbf{2.90} & \textbf{2.85} \\
        \hline
    \end{tabular}
\end{table}

\begin{figure}[!htb]
    \subfloat[]{\includegraphics[height=1.1in,trim={0.5cm 0.5cm 0.5cm 0.5cm},clip]{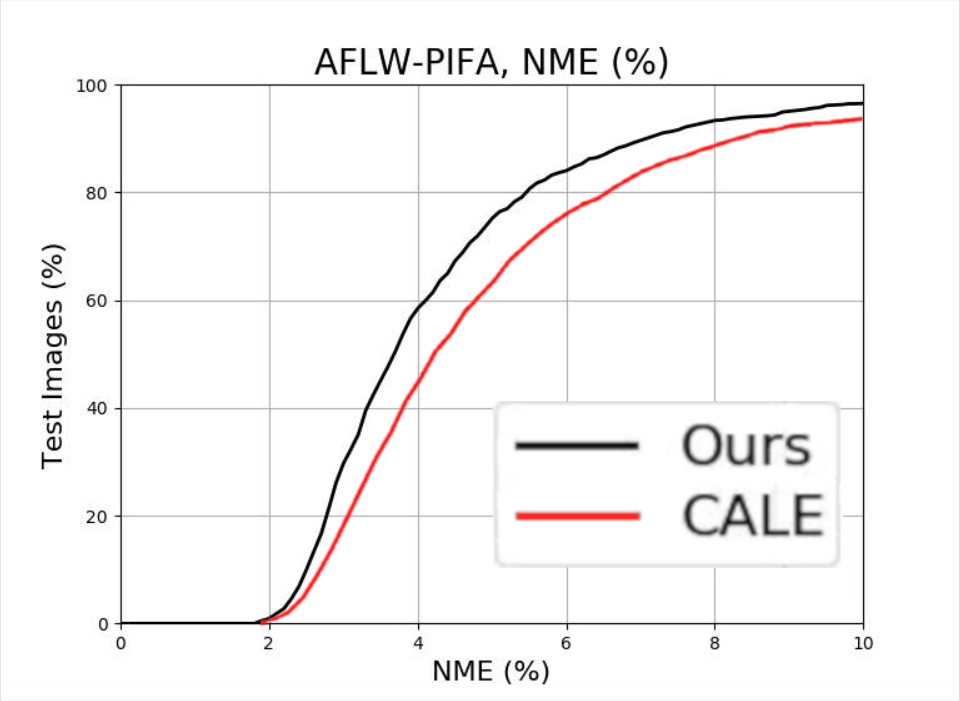}
        \label{fig:pifa_a}}
    \hfill
    \subfloat[]{\includegraphics[height=1.1in,trim={0.5cm 0.5cm 0.5cm 0.5cm},clip]{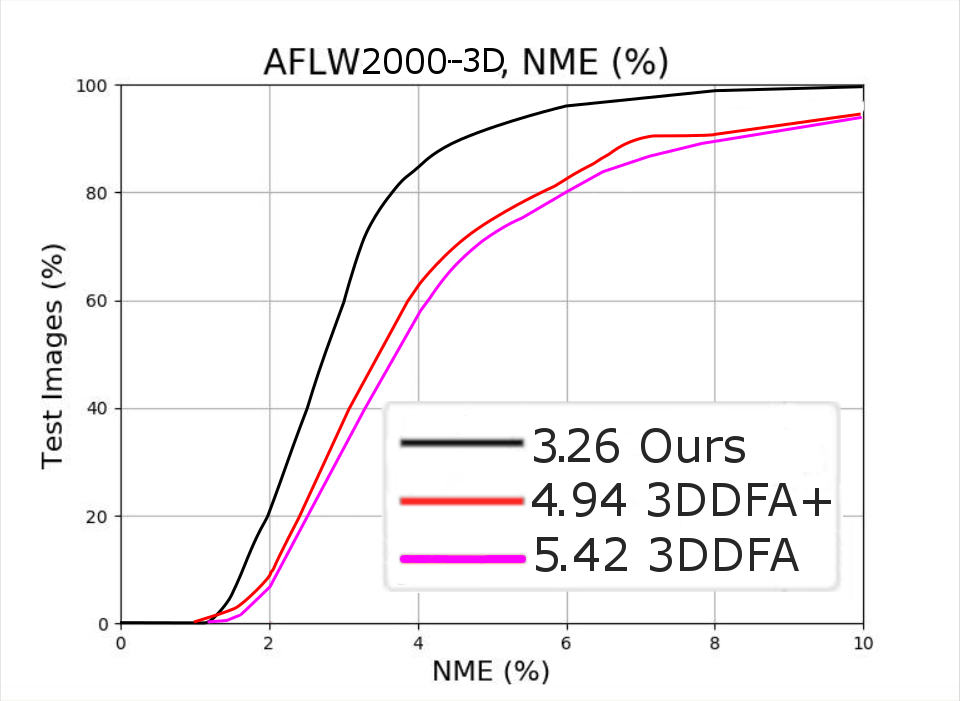}
        \label{fig:pifa_b}}
    \hfill
    \caption{Cumulative error curves (a) on AFLW-PIFA, evaluated on all 34 points (CALE is the method of \cite{bulat2016convolutional}), (b) on AFLW2000-3D on all points computed on a random subset of 696 images equally represented in $[0^\circ,30^\circ], [30^\circ,60^\circ], [60^\circ,90^\circ]$ (see also \cite{zhu2016face}).}
    \label{fig:pifa}
\end{figure}

AFLW-PIFA \cite{jourabloo2015pose} is a gray-scale subset of AFLW \cite{kostinger2011annotated}, consisting of 5,200 images (3,901 for training and 1,299 for testing) selected so that there is a balanced number of images for yaw angle in $[0^\circ,30^\circ], [30^\circ,60^\circ] \textrm{ and } [60^\circ,90^\circ]$. All images are annotated with 34 points from a 3D perspective. Fig.~\ref{fig:pifa_a} and Tables \ref{tab:pifa1} and \ref{tab:pifa2} show our results on AFLW-PIFA. When evaluated on both visible and occluded points, our method improves upon the current best result of \cite{bulat2016convolutional} (which uses real weights) by more than 10\%.

AFLW2000-3D is a subset of AFLW re-annotated by \cite{zhu2016face} from a 3D perspective with 68 points. We used this dataset only for evaluation. The training was done using the first 40,000 images from 300W-LP \cite{zhu2016face}. As Fig.~\ref{fig:pifa_b} shows, on AFLW2000-3D, the improvement over the  state-of-the-art method of \cite{zhu2016face} (real-valued) is even larger. As further results in Fig. \ref{tab:compreAFLW2000} show, while our method improves over the entire range of poses, the gain is noticeably higher for large poses ($[60^\circ-90^\circ]$), where we outperform \cite{zhu2016face} by more than 40\%.

\begin{table}[!htbp]
    \renewcommand{\arraystretch}{1.3}
    \caption{NME-based (\%) comparison on AFLW-PIFA evaluated on visible landmarks only. The results for PIFA, RCPR and PAWF are taken from \cite{jourabloo2016large}.}
    \label{tab:pifaall}
    \centering
    \begin{tabular}{|l|c|c|c|c|}
        \hline
        PIFA \cite{jourabloo2015pose} & RCPR \cite{burgos2013robust} & PAWF \cite{jourabloo2016large} & CALE \cite{bulat2016convolutional} & Ours \\
        \hline\hline
        8.04                          & 6.26                         & 4.72                           & \textbf{2.96}                      & 3.02 \\
        \hline
    \end{tabular} \label{tab:pifa1}
\end{table}

\begin{table}[!htbp]
    \renewcommand{\arraystretch}{1.3}
    \caption{NME-based (\%) based comparison on AFLW-PIFA evaluated on all 34 points, both visible and occluded. }
    \label{tab:pifa}
    \centering
    \begin{tabular}{|l|c|}
        \hline
        CALE \cite{bulat2016convolutional} & Ours          \\
        \hline\hline
        4.97                               & \textbf{4.47} \\
        \hline
    \end{tabular} \label{tab:pifa2}
\end{table}

\begin{table}[!htbp]
    \caption{NME-based (\%) based comparison on AFLW2000-3D evaluated on all 68 points, both visible and occluded. The results for RCPR, ESR and SDM are taken from \cite{zhu2016face}. }
    \label{tab:compreAFLW2000}
    \renewcommand{\arraystretch}{1.3}
    \centering
    \begin{tabular}{|l|c|c|c|c|}
        \hline
        Method                                  & [0,30]        & [30,60]       & [60,90]       & Mean          \\
        \hline\hline
        RCPR(300W) \cite{burgos2013robust}      & 4.16          & 9.88          & 22.58         & 12.21         \\
        RCPR(300W-LP) \cite{burgos2013robust}   & 4.26          & 5.96          & 13.18         & 7.80          \\
        ESR(300W) \cite{cao2014face}            & 4.38          & 10.47         & 20.31         & 11.72         \\
        ESR(300W-LP) \cite{cao2014face}         & 4.60          & 6.70          & 12.67         & 7.99          \\
        SDM(300W) \cite{xiong2013supervised}    & 3.56          & 7.08          & 17.48         & 9.37          \\
        SDM(300W-LP) \cite{xiong2013supervised} & 3.67          & 4.94          & 9.76          & 6.12          \\
        3DDFA \cite{zhu2016face}                & 3.78          & 4.54          & 7.93          & 5.42          \\
        3DDFA+SDM \cite{zhu2016face}            & 3.43          & 4.24          & 7.17          & 4.94          \\
        \hline
        \textbf{Ours}                           & \textbf{2.47} & \textbf{3.01} & \textbf{4.31} & \textbf{3.26} \\
        \hline
    \end{tabular} 
\end{table}

\begin{figure*}[!htb]
    \centering
    \subfloat[\textbf{(Ours, final)} binary block with varying depth. See also Subsection~\ref{ssec:conv-depth}.]{\makebox[0.45\textwidth]{\includegraphics[height=2.5in,trim={0.5cm 0.5cm 0.5cm 0.5cm},clip]{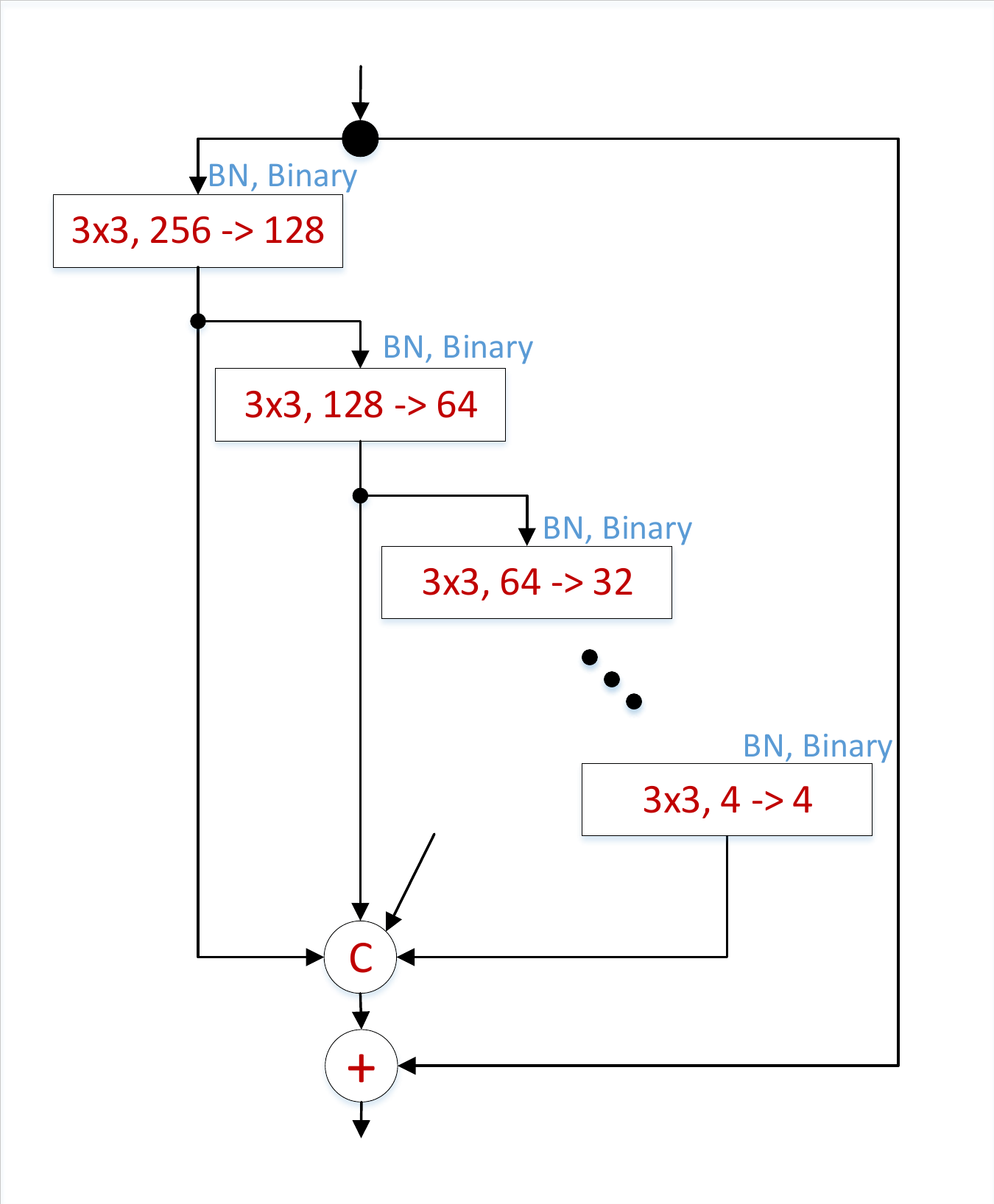}}
    \label{fig:layers_ours_deep}}
    \hfill
    \subfloat[Depth vs PCKh-based performance on the MPII validation set.]{\begin{tikzpicture}[object/.style={thin,double,<->},scale=0.85]
        \begin{axis}[
            title={Performance vs block depth},
            xlabel={Depth (number of layers)},
            ylabel={PCKh, MPII validation set (\%)},
            xmin=3, xmax=8,
            ymin=74, ymax=77,
            xtick={3,4,5,6,7,8},
            ytick={74.0, 74.5,75.0,75.5,76,76.5,77},
            legend pos=south east,
            ymajorgrids=true,
            xmajorgrids=true,
            grid style=dashed,
        ]
        
        \addplot[
            mark max,
            color=blue,
            mark=*,
            smooth,
            tension=1
            ]
            coordinates {
            (3,76.0)(4,76.2)(5,76.4)(6,76.5)(7,76.4)(8,76.3)
            };
            \legend{}
        \end{axis}
    \end{tikzpicture}\label{fig:depth-relation}}
    \caption{The effect of varying the depth of the proposed binary block on performance.}
    \label{fig:depth-both}
\end{figure*}
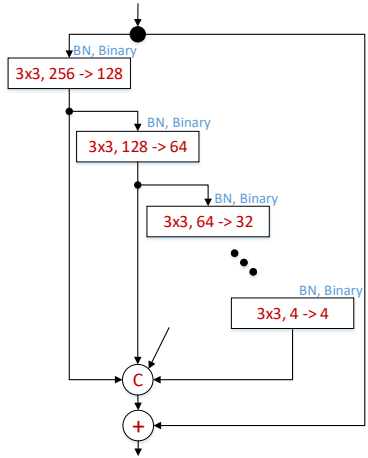
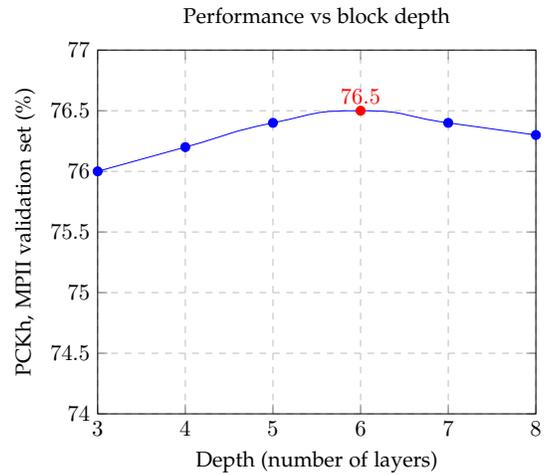

\section{Advanced block architectures}\label{sec:advanced-block-architectures}

In this section, we explore the effectiveness of two architectural changes applied to our best performing block (\textbf{Ours, final}), namely varying its depth and its cardinality. Again, we used the standard training-validation partition of MPII.

\subsection{On the depth of the proposed block}\label{ssec:conv-depth}

To further explore the importance of the multi-scale component in the overall structure of the proposed block, we gradually increase its depth and as a result, the number of its layers, as shown in  Fig.~\ref{fig:depth-relation}. The advantage of doing this is twofold: (a) it increases the receptive field within the block, and (b) it analyses the input simultaneously at multiple scales. We ensure that by doing so the number of parameters remains (approximately) constant. To this end, we halve the number of channels of the last layer at each stage. In the most extreme case, the last layer will have a single channel. Because, the representational power of such a small layer is insignificant, in practice we stop at a minimum of 4, which corresponds to a depth equal to 8. The results, reported in Fig.~\ref{fig:depth-relation}, show that the  performance gradually improves up to 76.5\% for a depth equal to 6, and then, further on, it saturates and eventually gradually degrades as the depth increases. 

\textbf{Conclusion:} The depth of the multi-scale component is an important factor on the overall module performance. Increasing it, up to a certain point, is beneficial and can further improve the performance at no additional cost.


\subsection{On the cardinality of the proposed block}\label{ssec:conv-cardinality}

Inspired by the recent innovations of \cite{xie2016aggregated} for real-valued networks, in this section we explore the behavior of an increased cardinality (defined as in \cite{xie2016aggregated} as the size of the set of transformations) when applied to our binary hierarchical, parallel \& multi-scale block.

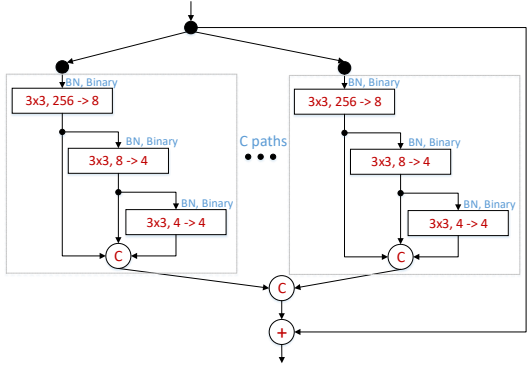
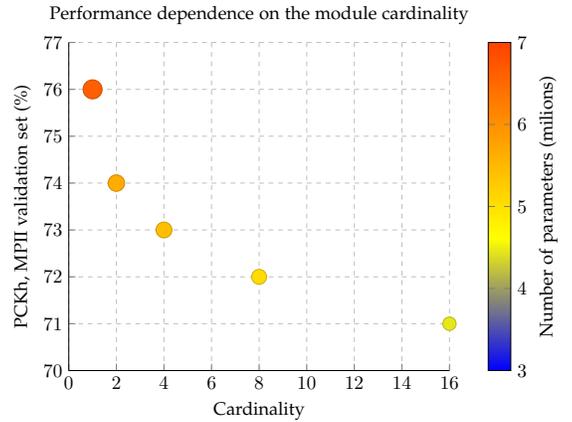
\begin{figure*}[!htb]
    \centering
    \subfloat[ResNetXt-like extension of \textbf{(Ours, final)} binary block. C represents the cardinality of the block. See also Subsection~\ref{ssec:conv-cardinality}.]{\makebox[0.45\textwidth]{\includegraphics[height=2.0in,trim={0.5cm 0.5cm 0.5cm 0.5cm},clip]{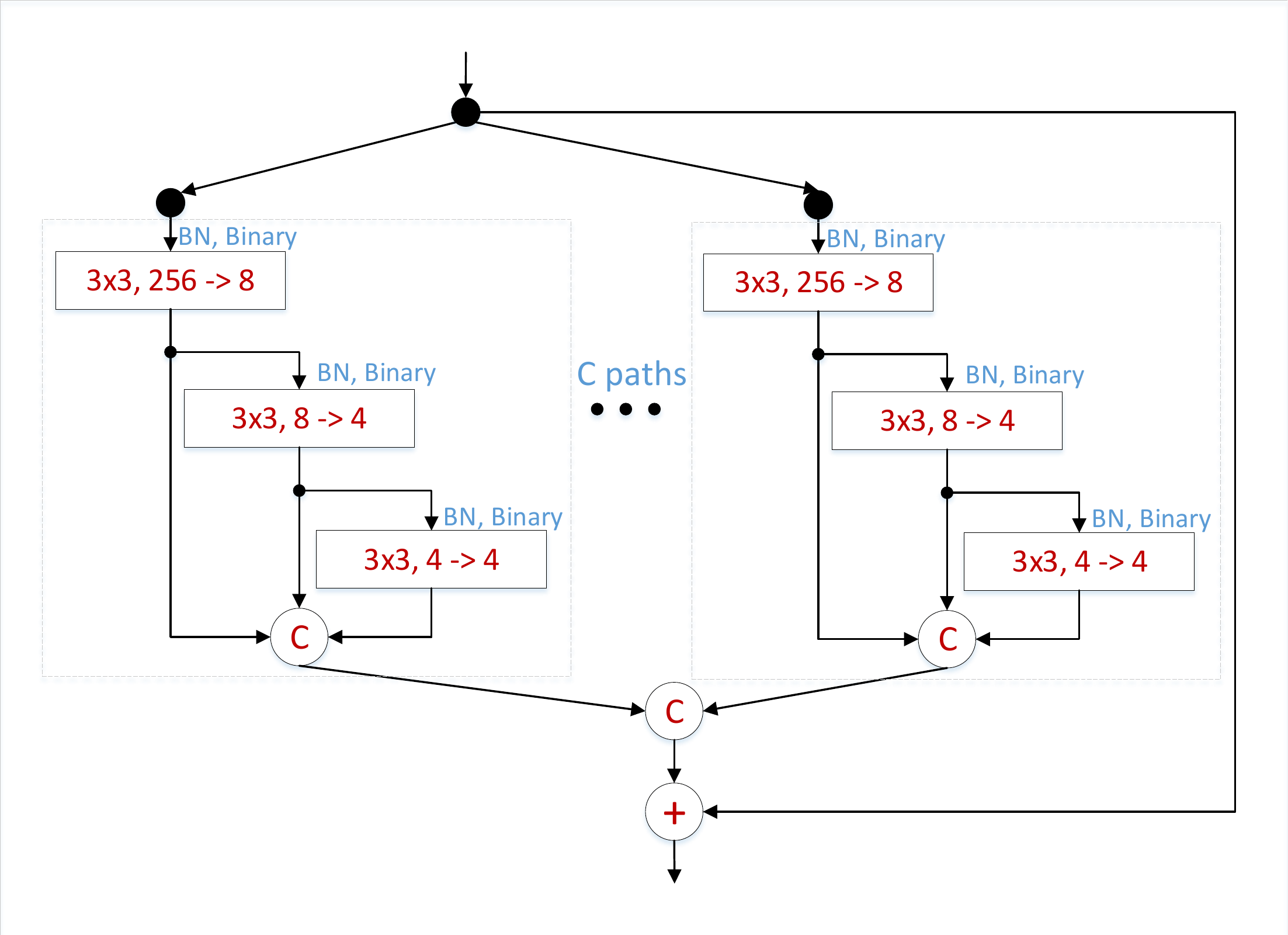}}\label{fig:layers_resnext}}
    \hfill
    \subfloat[Cardinality vs PCKh-based performance on the MPII validation set. Notice how the efficiency (the ratio between the number of parameters and PCKh) decreases as we increase the block cardinality.]{\begin{tikzpicture}[object/.style={thin,double,<->},scale=0.6, every node/.style={scale=1.2}]
        \begin{axis}[%
            title={Performance dependence on the module cardinality},
            xlabel={Cardinality}, 
            ylabel={PCKh, MPII validation set (\%)},
            zlabel={},
            scale only axis,
            xmin=0, xmax=16,
            ymin=70, ymax=77,
            axis x line*=bottom,
            axis y line*=left,
            ymajorgrids=true,
            xmajorgrids=true,
            grid style=dashed,
            point meta min=3,
            point meta max=7,
            colorbar,
            colorbar style={
                ylabel={Number of parameters (milions)}
            }
        ]
        
        \addplot[%
            scatter,
            only marks,
            scatter src=explicit,
            mark=*,
            color=blue,
            visualization depends on = {\thisrow{C} \as \perpointmarkersize},
            scatter/@pre marker code/.append style={/tikz/mark size=\perpointmarkersize},
        ] table [meta=C] {temp.dat};
        \end{axis}
        \end{tikzpicture}%
        \label{fig:cardinality-realtion}}
        \caption{The effect of varying the cardinality of the proposed binary block on performance.}
        \label{fig:cardinality-all}
\end{figure*}

Starting again from our block of Fig.~\ref{fig:layers_parallel}, we replicate its structure $C$ times making the following adjustments in the process: (1) While the number of input channels of the first layer remains the same, the output and the input of the subsequent layers are reduced by a factor of $C$, and (2) the output of the replicated blocks is recombined via concatenation. The final module structure is depicted in Fig.~\ref{fig:cardinality-realtion}.

The full results with respect to the network size and the block cardinality (ranging from 1 to 16) are shown in Fig.~\ref{fig:cardinality-realtion}. Our findings are that increasing the block cardinality, while shown to provide good improvement on image classification using real-value networks, for the case of binary networks, given their significantly smaller size, depth and representational power, the same observation does not hold. In particular, when incorporated into the structure of our block with a similar number of parameters, the module under-performs by 1\% compared to the original block (having a cardinality equal to one). \newline
\textbf{Conclusion:} For the binary case, further increasing the block cardinality hurts performance.

\section{Improved network architectures} \label{sec:network-architectures}

In all previous sections, we investigated the performance of the various blocks by incorporating them into a single hourglass network, i.e. by keeping the network architecture fixed. In this section, we explore a series of architectural changes applied to the overall network structure. First, inspired by \cite{ronneberger2015u}, we simplify the HG model, improving its performance without sacrificing accuracy for the binary case. Then, we study the effect of stacking multiple networks together and analyze their behavior.

\begin{figure}[!htb]
    \centering
    \includegraphics[height=1.4in,trim={0cm 0cm 0cm 0cm},clip]{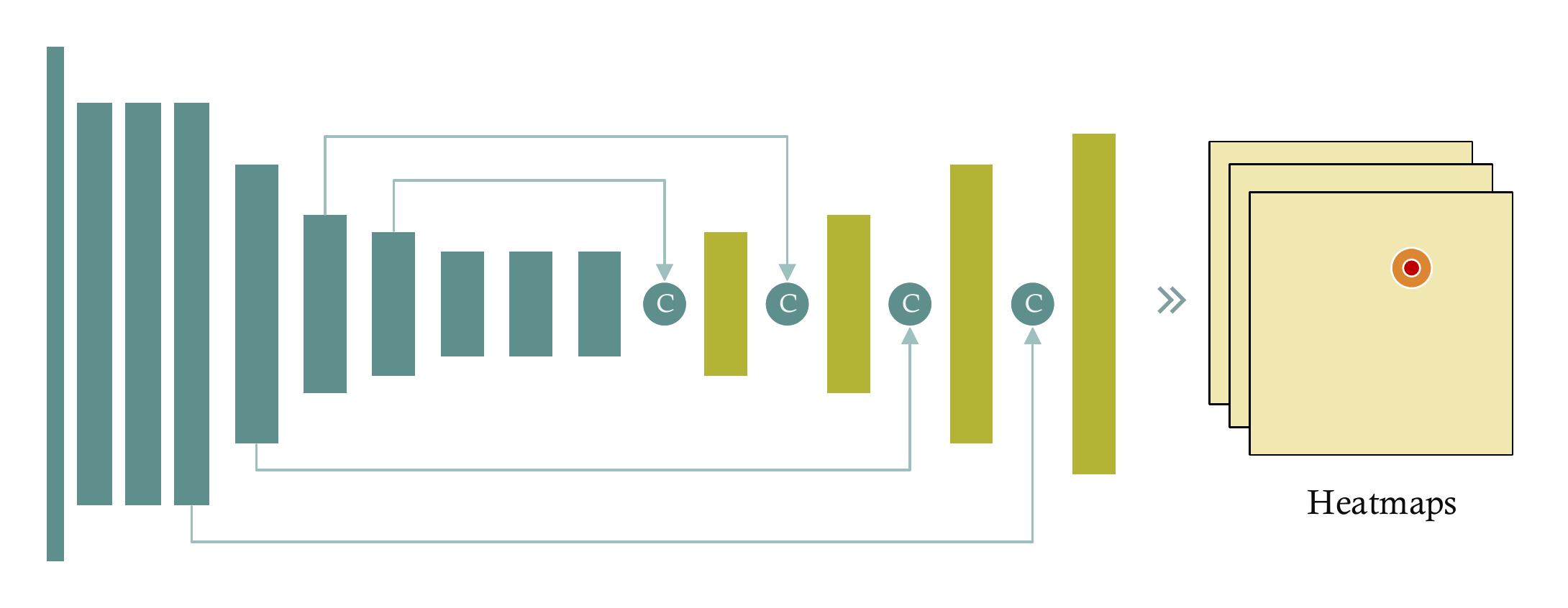}
    \caption{Improved, U-Net inspired, HG architecture. The dark-green modules were left unchanged, while for the light-green ones we doubled the number of their input channels from 256 to 512.}
    \label{fig:improved-hg}
\end{figure}

\subsection{Improved HG architecture}\label{ssec:improved-hg}
Motivated by the findings of Subsection~\ref{ssec:better} that shed light on the importance of the gradient flow and suggested that skip connections with shorter paths should be used where possible, we adopt a similar approach to the overall HG architecture.

In particular, to improve the overall gradient flow, we removed the residual blocks in the upsampling branches that are tasked with the ``injection'' of high resolution information into the later stages of the network. To adjust to that change, the number of input channels of the first layer from the modules that are immediately after the point where the branch is merged via \textit{concatenation} is increased by two times (to accommodate to the increase in the number of channels). {\color{black} The resulting architecture, depicted in Fig.~\ref{fig:improved-hg}, is a modified U-net architecture\cite{ronneberger2015u} which was binarized in the same way  as the HG model.}

The results, reported in Table~\ref{tab:improved_hg}, show that by removing the residual blocks from the upsampling branches, the performance, over the baseline HG is increased by 0.5\%, further solidifying the importance of the gradient flow in the performance of binary networks. Furthermore, due to the decrease in the number of layers and parameters, an up to 20\% speedup is observed.
The network is trained using the same procedure described previously, for 100 epochs.

\begin{table}[!htbp]
    \renewcommand{\arraystretch}{1.3}
    \caption{Comparison between HG and Improved HG on the MPII validation set. Both networks are built with our proposed binarized block.}
    \label{tab:improved_hg}
    \centering
    \begin{tabular}{|l|c|c|}
        \hline
        Network architecture                    & \# parameters & PCKh            \\
        \hline\hline
        HG (Fig.~\ref{fig:network_small})                & 6.2M          & 76\%            \\
        \hline
        \textbf{Improved HG} (Fig.~\ref{fig:improved-hg}) & \textbf{5.8M} & \textbf{76.6\%} \\
        \hline
    \end{tabular}
\end{table}

\subsection{Stacked Binarized HG networks}\label{ssec:stack-hg}

\begin{figure}[!htb]
    \centering
    \includegraphics[height=1.4in,trim={0cm 0cm 0cm 0cm},clip]{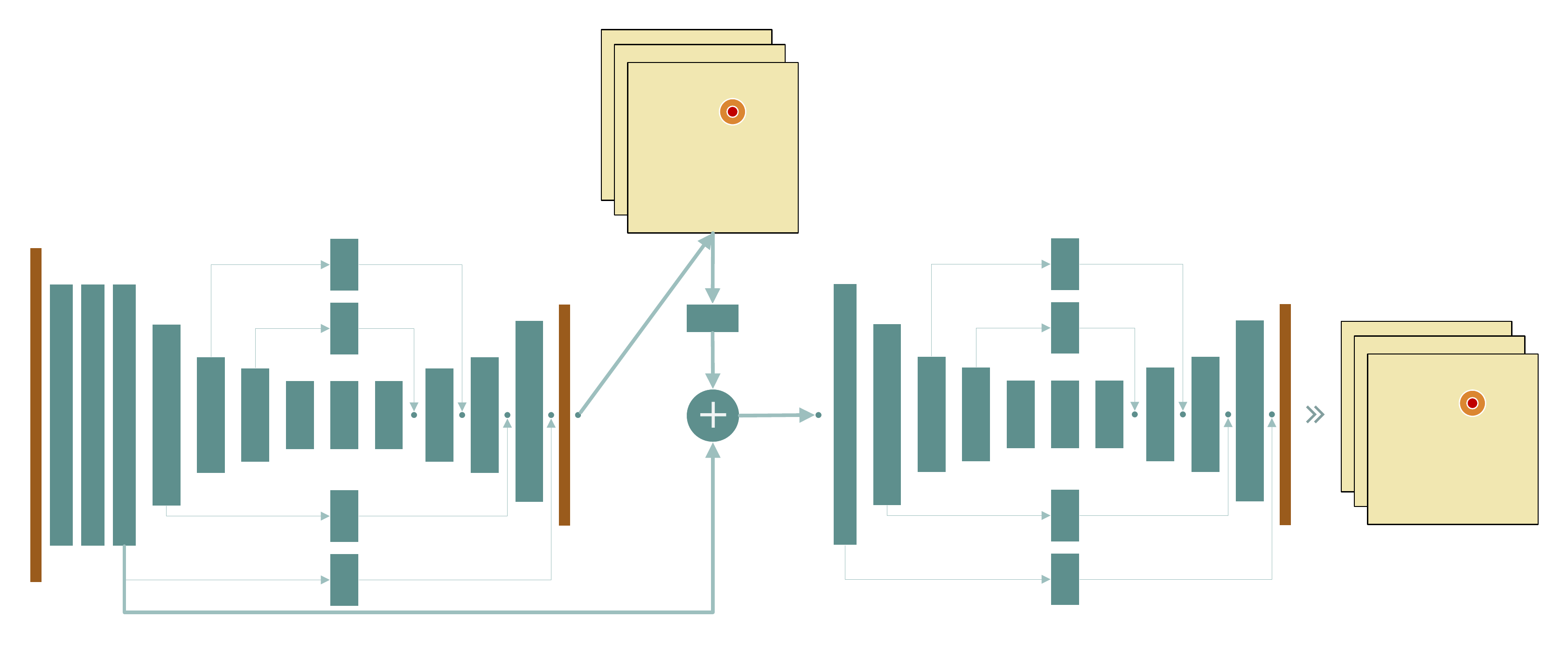}
    \caption{A two-stack binarized HG. All blocks are binarized, except for the very first and last layers showed in red colour.}
    \label{fig:stacked-hg}
\end{figure}

\begin{figure}[!htb]
    \centering
    \includegraphics[height=2.9in,trim={0.5cm 0.5cm 0.5cm 0.5cm},clip]{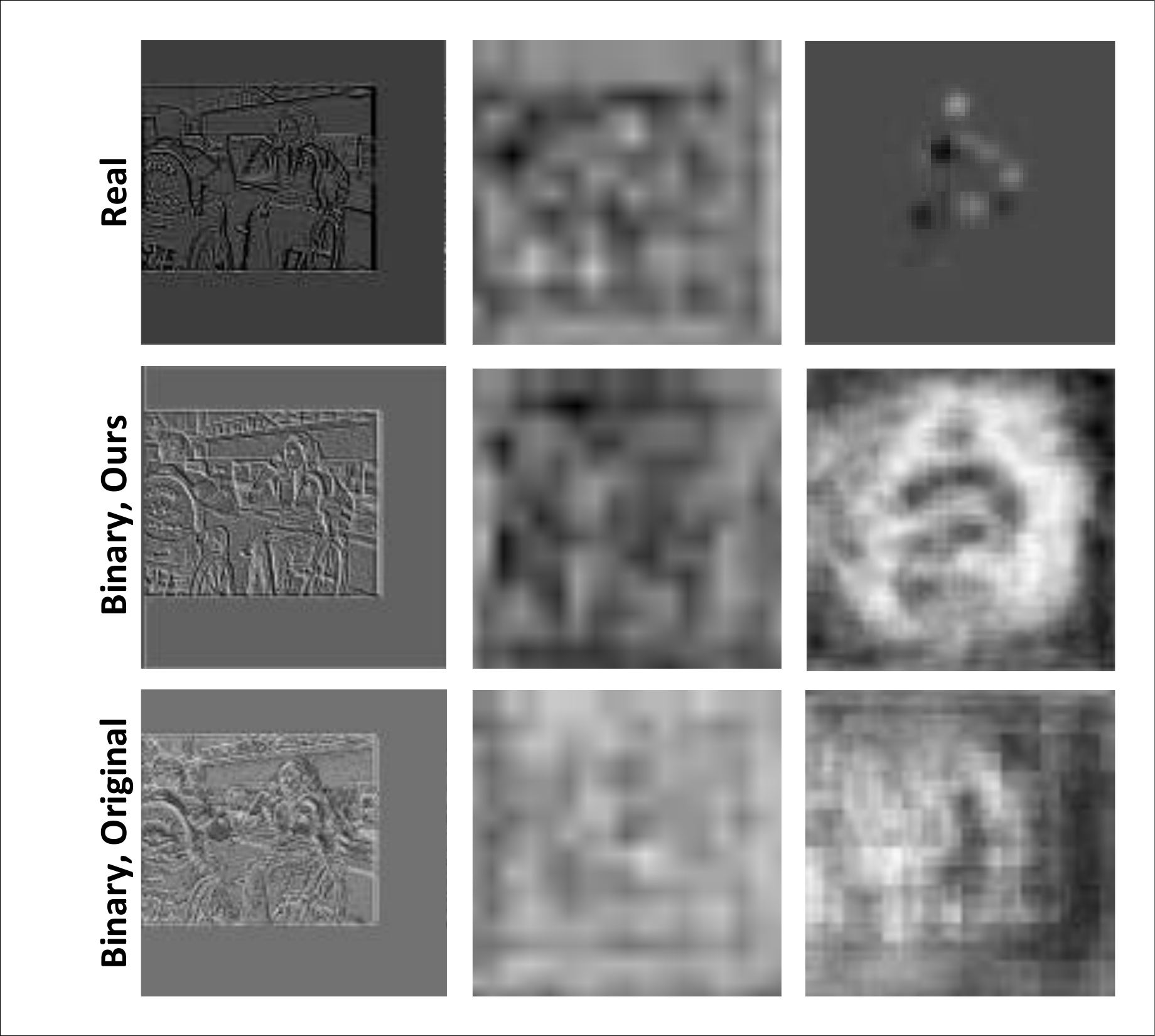}
    \caption{Features extracted from the first layer (first column), middle layer (middle column) and right before the very last layer (right column) for real-valued and binary (ours and original) networks. As we move on to the last layers, activations become more noisy for the binary case, which we believe that it hurts the performance of the stacked networks.}
    \label{fig:activation_maps}
\end{figure}

Network stacking was recently shown to achieve state-of-the-art results on human pose estimation\cite{bulat2016human,newell2016stacked,wei2016convolutional} when real-valued models are used. In this subsection, we explore whether the same holds for the binary case.

Following \cite{newell2016stacked}, we stack and interconnect the networks as follows: The first network takes as input the RGB image and outputs a set of $N$ heatmaps. The next network in the stack takes as input the sum of: (1) the input to the previous network, (2) the projection of the previously  predicted heatmaps, and (3) the output of the last but one block from the previous level. The resulting network for a stack of two is shown in Fig.~\ref{fig:stacked-hg}. 

As the results of Table~\ref{tab:stacked_hg} show, network stacking  for the binary case behaves to some extent similarly to the real-valued case, however the gains from one stage to another are smaller, and performance seems to saturate faster. We believe that the main reason for this is that for the case of binary networks, activations are noisier especially for the last layers of the network. This is illustrated in Fig. \ref{fig:activation_maps} where we compare the feature maps obtained from a real and the two types of binary networks compared in this paper (original, based on bottleneck and proposed). Clearly the feature maps for the binary case are more noisy and blurry as we move on to the last layers of the network. As network stacking relies on features from the earlier networks of the cascade and as these are noisy, we conclude that this has a negative impact on the overall network's performance.   

\begin{table}[!htbp]
    \renewcommand{\arraystretch}{1.3}
    \caption{Accuracy of stacked networks on MPII validation set. All networks are built with our proposed binarized block.}
    \label{tab:stacked_hg}
    \centering
    \begin{tabular}{|l|c|c|}
        \hline
        \# stacks   & \# parameters & PCKh            \\
        \hline\hline
        1                & 6.2M          & 76\%            \\
        \hline
        2 & 11.0M & 79.9\% \\
        \hline
        3 & 17.8M & 81.3\% \\
        \hline
    \end{tabular}
\end{table}

\textbf{Training.} To speedup the training process, we trained the stacked version in a sequential manner. First, we trained the first network until convergence, then we added the second one on top of it, freezing its weights and training the second one. The process is repeated until all networks are added. Finally, the entire stack is trained jointly for 50 epochs. 

\section{Additional experiments}\label{sec:additional-experiments}

\begin{figure}[!htb]
    \centering
    \includegraphics[height=1.35in, trim={0.5cm 0.5cm 0.5cm 0.5cm},clip]{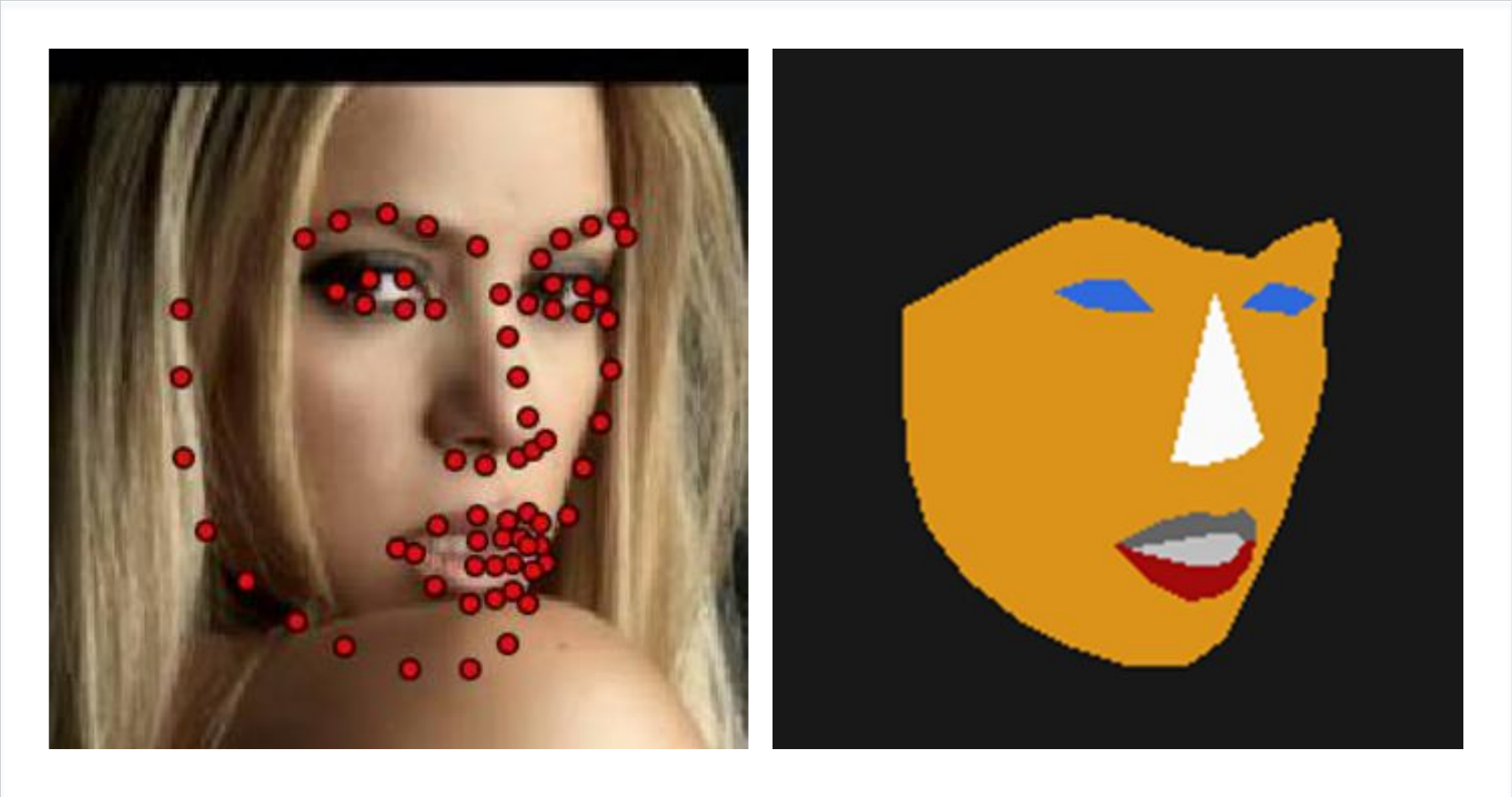}
    \caption{Example of a ground truth mask (right) produced by joining the 68 ground truth landmarks (left). Each colour denotes one of the seven classes.}
    \label{fig:seg_gt_example}
\end{figure}

In this section, we further show that the proposed block generalizes well producing consistent results across various datasets and tasks. To this end, we report results on the task of face parsing, also known as semantic facial part segmentation, which is the problem of assigning a categorical label to every pixel in a facial image. We constructed a dataset for facial part segmentation by joining together the 68 ground truth landmarks (originally provided for face alignment) to fully enclose each facial component. In total, we created seven classes: skin, lower lip, upper lip, inner mouth, eyes, nose and background. Fig.{~\ref{fig:seg_gt_example}} shows an example of a ground truth  mask. We trained the network on the 300W dataset (approximately 3,000 images) and tested it on the 300W competition test set, both Indoor\&Outdoor subsets (600 images), using the same procedure described in Section 7.

\textbf{Architecture}. We reused the same architecture for landmark localization, changing only the last layer in order to accommodate the different number of output channels (from 68 to 7). We report results for three different networks of interest: (a) a real-valued network using the original bottleneck block (called ``Real, Bottleneck''), (b) a binary network using the original bottleneck block (called ``Binary, Bottleneck''), and (c) a binary network using the proposed block (called ``Binary, Ours''). To allow for a fair comparison, all  networks have a similar number of parameters and depth. For training the networks, we used the Log-Softmax loss {\cite{long2015fully}}.

\begin{figure*}[!t]
    \centering
    \includegraphics[height=5.0in,trim={0.5cm 0.0cm 0.5cm 0.5cm},clip]{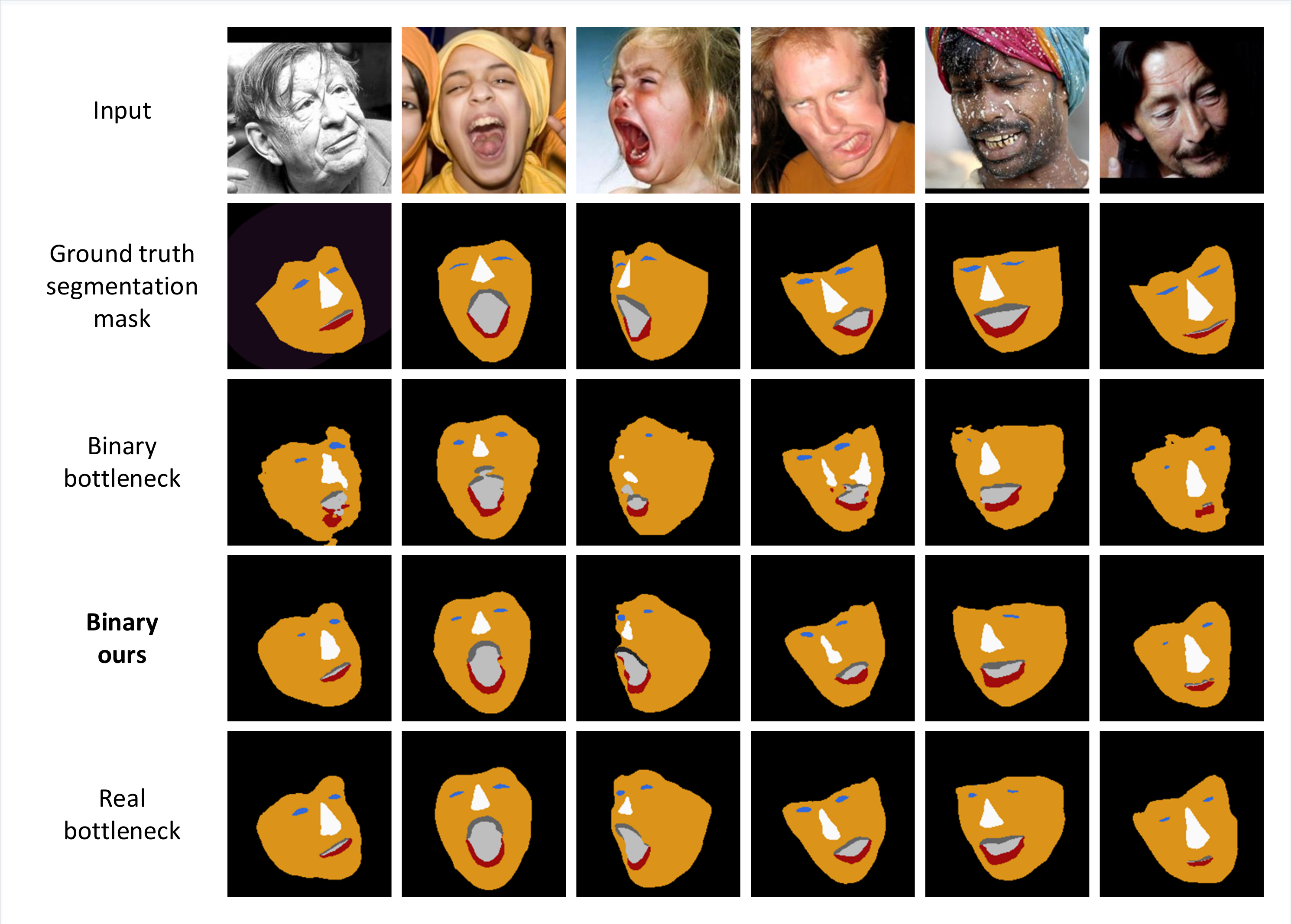}
    \caption{Qualitative results on 300W (Indoor\&Outdoor). Observe that the proposed binarized network significantly outperforms the original binary one, almost matching the performance of the real-valued network.}
    \label{fig:examples_segm}
\end{figure*}

\textbf{Results.}  Table{~\ref{tab:segm_results}} shows the obtained results. Similarly to our human pose estimation and face alignment experiments, we observe that the binarized network based on the proposed block significantly outperforms a similar-sized network constructed using the original bottleneck block, almost matching the performance of the real-valued network. Most of the performance improvement is due to the higher representation/learning capacity of our block, which is particularly evident for difficult cases like unusual poses, occlusions or challenging lighting conditions. For visual comparison, see  Fig.{~\ref{fig:examples_segm}}.
\begin{table}[!htbp]
    \renewcommand{\arraystretch}{1.3}
    \caption{Results on 300W (Indoor\&Outdoor). The pixel acc., mean acc. and mean IU are computed as in {\cite{long2015fully}}.}
    \label{tab:segm_results}
    \centering
    \begin{tabular}{|l|c|c|c|}
        \hline
        Network type          & pixel acc. & mean acc. & mean IU \\
        \hline\hline
        Real, bottleneck      & 97.98\%    & 77.23\%   & 69.29\% \\
        Binary, bottleneck    & 97.41\%    & 70.35\%   & 62.49\% \\
        \textbf{Binary, Ours} & 97.91\%    & 76.02\%   & 68.05\% \\
        \hline
    \end{tabular}
\end{table}

\begin{figure*}[!t]
    \centering
    \subfloat[Fitting examples produced by our binary network on AFLW2000-3D dataset. Notice that our method copes well with extreme poses, facial expressions and lighting conditions.]{\includegraphics[height=2.2in,trim={0.5cm 0.5cm 0.5cm 0.5cm},clip]{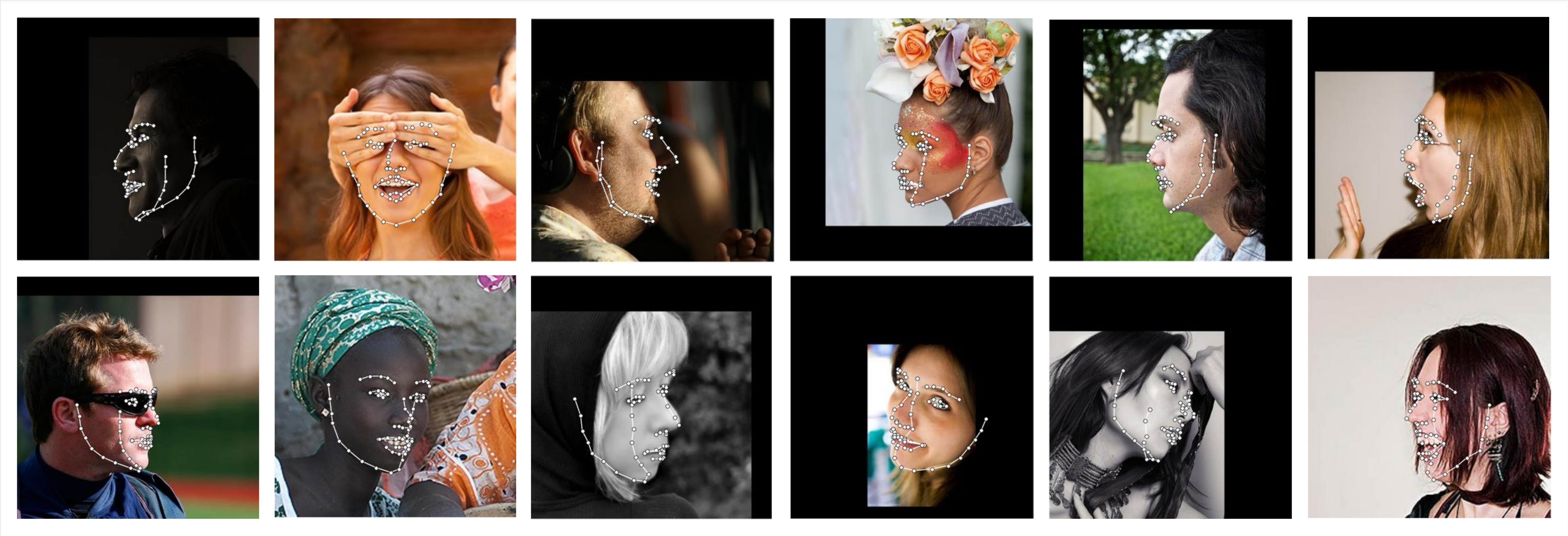}
        \label{fig:examples_face}}

    \subfloat[Examples of human poses obtained using our binary network. Observe that our method produces good results for a wide variety of poses and occlusions.]{\includegraphics[height=2.2in,trim={0.5cm 0.5cm 0.5cm 0.5cm},clip]{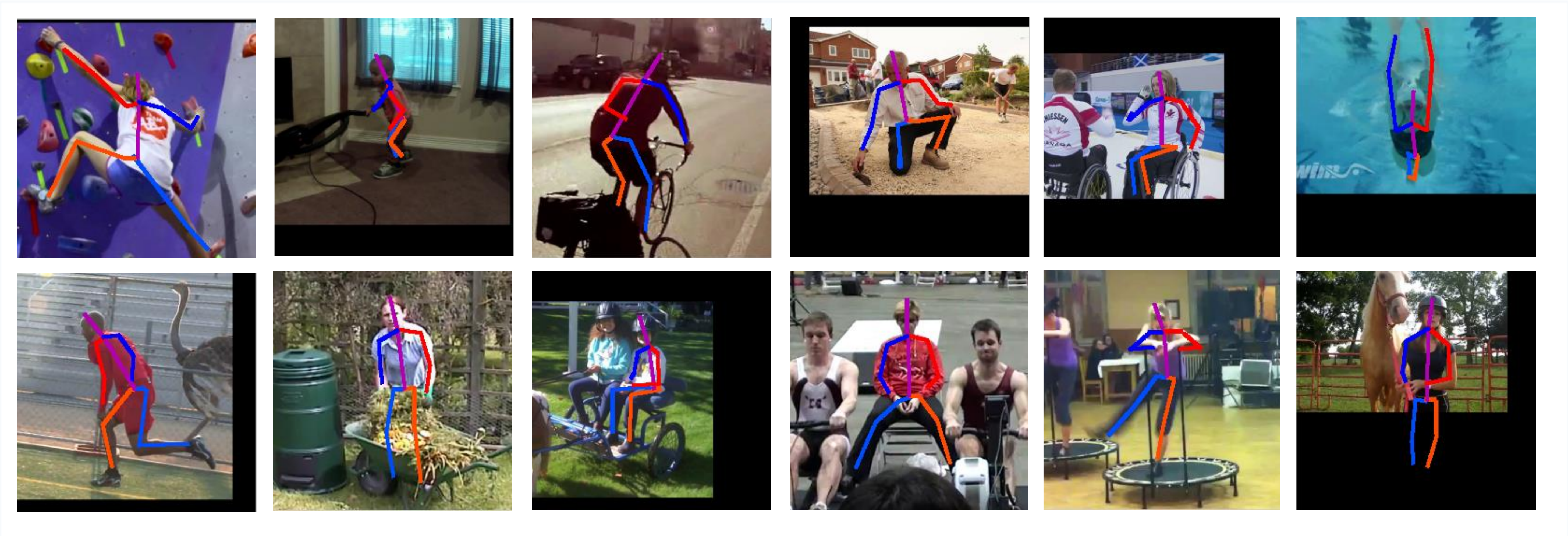}
		\label{fig:examples_human}}
		
    \caption{Qualitative results produced by our method on (a) AFLW2000-3D and (b) MPII datasets.}
    \label{fig:examples}
\end{figure*}

\section{Conclusion} \label{sec:conclusion}
We proposed a novel block architecture, particularly tailored for binarized CNNs and localization visual tasks. During the process, we exhaustively evaluated various design choices, identified performance bottlenecks and proposed solutions. We showed that our hierarchical, parallel and multi-scale block enhances representational power, allowing for stronger relations to be learned without excessively increasing the number of network parameters. The proposed architecture is efficient and can run on limited resources. We verified the effectiveness of the proposed block on a wide range of fine-grained recognition tasks including human pose estimation, face alignment, and facial part segmentation.



\ifCLASSOPTIONcompsoc
    \section*{Acknowledgments}
\else
    \section*{Acknowledgment}
\fi

Adrian Bulat was funded by a PhD scholarship from the University of Nottingham. This work was supported by the Engineering and Physical Sciences Research Council [grant number EP/M02153X/1] Facial Deformable Models of Animals to Georgios Tzimiropoulos.

\ifCLASSOPTIONcaptionsoff
\newpage
\fi



%

{
    \bibliographystyle{IEEEtran}
    \bibliography{egbib}
}



%
\vfill

\begin{IEEEbiography}[{\includegraphics[height=1.25in,clip,keepaspectratio]{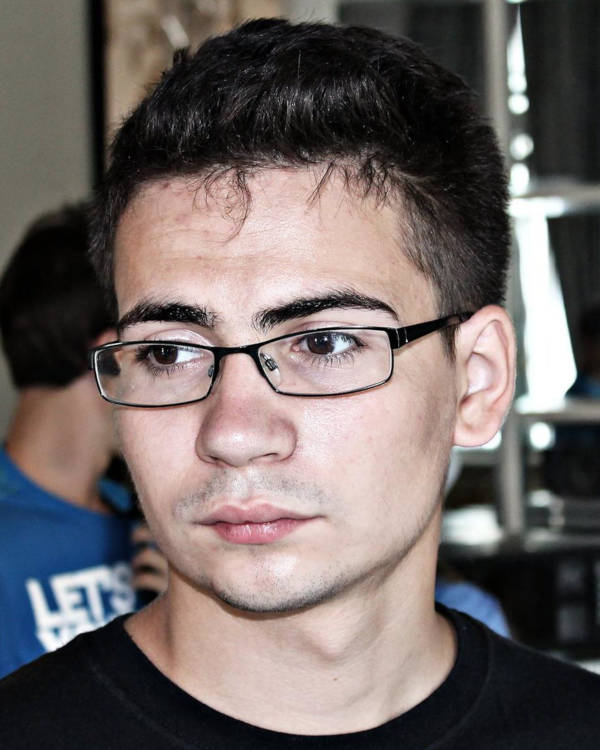}}]{Adrian Bulat}
    is currently a PhD student with the Computer Vision Laboratory at the University of Nottingham, under the supervision of Dr. Georgios Tzimiropoulos. He received his B.Eng. in Computer Engineering (2015) from the Technical University ``Gheorghe Asachi'' (Romania).
\end{IEEEbiography}

\begin{IEEEbiography}[{\includegraphics[height=1.25in,clip,keepaspectratio]{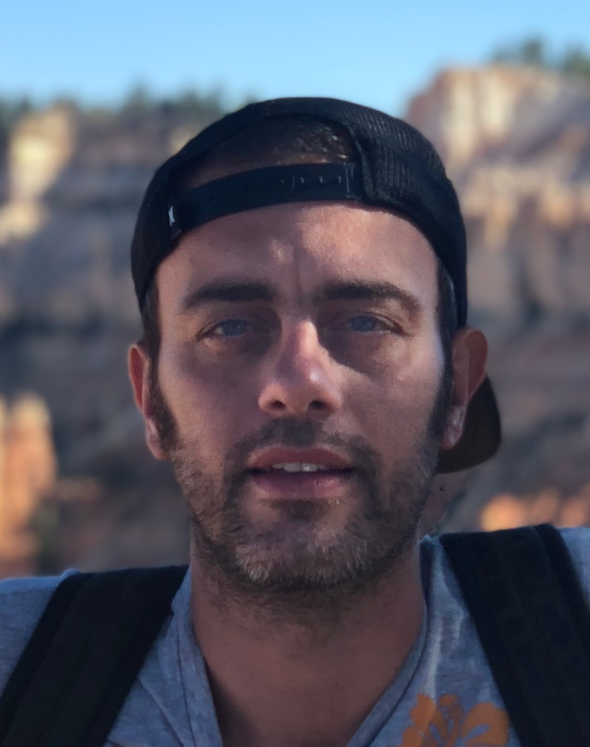}}]{Georgios (Yorgos) Tzimiropoulos}  received the M.Sc. and Ph.D. degrees in Signal Processing and Computer Vision from Imperial College London, U.K. He is Assistant Professor with the School of Computer Science at the University of Nottingham, U.K. Prior to this, he was a Senior Researcher in the iBUG group, Department of Computing, Imperial College London. He is currently Associate Editor of the Image and Vision Computing Journal. He has worked on the problems of object detection and tracking, alignment and pose estimation, and recognition with humans and faces being the focal point of his research. For his work, he has used a variety of tools from Mathematical Optimization and Machine Learning. His current focus is on Deep Learning. 
\end{IEEEbiography}






\end{document}